\documentclass[runningheads]{llncs}

\usepackage{style}

\usepackage{styleabbrv}

\usepackage{graphicx}
\usepackage{booktabs}

\usepackage{adjustbox}
\usepackage{caption}
\usepackage{array}
\usepackage[table]{xcolor}

\usepackage{placeins}
\usepackage{comment}

\usepackage{listings}
\usepackage{fontawesome5}
\usepackage[breakable,skins]{tcolorbox}

\usepackage[english]{babel}

\usepackage{amsmath,amsfonts,graphicx}

\usepackage{microtype}
\usepackage{cite}

\usepackage{tabularx}
\usepackage{multirow}

\usepackage{wrapfig}

\usepackage{hyperref}
\hypersetup{
  breaklinks=true,
  colorlinks=true,
  linkcolor=blue,
  citecolor=blue,
  urlcolor=black,
  hypertexnames=false,
  bookmarksnumbered=true,
  pdftitle={Embed-RL: Reinforcement Learning for Reasoning-Driven Multimodal Embeddings},
  pdfauthor={Haonan Jiang et al.}
}

\usepackage[accsupp]{axessibility}

\usepackage{orcidlink}

\begin{document}

\title{Embed-RL: Reinforcement Learning for Reasoning-Driven Multimodal Embeddings}

\titlerunning{Embed-RL: RL for Reasoning-Driven Multimodal Embeddings}

\author{Haonan Jiang\inst{1,2*}\and
Yuji Wang\inst{1,2*} \and
Yongjie Zhu\inst{2 \dag}\and
Xin Lu\inst{2}\and
Wenyu Qin\inst{2}\and
Meng Wang\inst{2}\and
Pengfei Wan\inst{2} \and
Yansong Tang\inst{1 \ddag}}

\authorrunning{H.~Jiang, Y.~Wang et al.}

\institute{
$^1$Tsinghua Shenzhen International Graduate School, Tsinghua University \\
$^2$Kling Team, Kuaishou Technology\\
\email{\{jiang-hn24@mails, yuji-wan24@mails, tang.yansong@sz\}.tsinghua.edu.cn} \\
\email{\{zhuyongjie, luxin09, qinwenyu, wangmeng46, wanpengfei\}@kuaishou.com}
}

\maketitle

\let\thefootnote\relax\footnotetext{$*$: Equal Contribution. Work done during an internship at Kuaishou Technology. \\ $\dag$: Project Leader. $\ddag$: Corresponding Author.}

\begin{abstract}
Leveraging Multimodal Large Language Models (MLLMs) has become pivotal for advancing Universal Multimodal Embeddings (UME) in addressing diverse cross-modal tasks. Recent studies demonstrate that incorporating generative Chain-of-Thought (CoT) reasoning can substantially enhance task-specific representations compared to discriminative methods. However, the generated reasoning CoTs of existing generative embedding methods are limited to the textual analysis of queries and are irrelevant to the retrieval of the targets. To address these limitations, we propose a reasoning-driven UME framework that integrates Embedder-Guided Reinforcement Learning (EG-RL) to optimize the Reasoner to produce evidential Traceability CoT (T-CoT). 
Our key contributions are threefold: 
(1) We design an EG-RL framework where the Embedder provides explicit supervision to the Reasoner, ensuring the generated CoT traces are aligned with embedding tasks.
(2) We introduce T-CoT, which extracts critical multimodal cues to focus on retrieval-relevant elements and provides multimodal inputs for the Embedder. 
(3) With limited computational resources, our framework outperforms the pioneering embedding model on both MMEB-V2 and UVRB benchmarks. 
The integration of multimodal evidence in structured reasoning, paired with retrieval-oriented alignment, effectively strengthens cross-modal semantic consistency and boosts the model’s fine-grained matching capability as well as its generalization across complex scenarios.
Our work demonstrates that targeted reasoning optimization can significantly improve multimodal embedding quality, providing a practical and efficient solution for reasoning-driven UME development.
\href{https://github.com/ZoengHN/Embed-RL}{Project page.}

\keywords{Multimodal Embedding \and Generative Reasoning \and Reinforcement Learning}
\end{abstract}
\section{Introduction}
\label{sec:intro}

Multimodal embedding, as a core supporting technology for cross-modal tasks, has been widely applied to numerous important directions such as image-text retrieval, video moment localization, and visual document understanding \cite{jiang2024vlm2vec, lin2024mm}. Traditional multimodal embedding methods adopt dual-encoder architectures, such as CLIP \cite{radford2021learning}, BLIP \cite{li2023blip}, and SigLIP \cite{zhai2023sigmoid}. These methods demonstrate weaker ability in bridging the gap between different modalities compared with Multimodal Large Language Models (MLLMs) \cite{li2024llava, li2024llava-next, wang2024qwen2, bai2025qwen2, Qwen3-VL}. Additionally, MLLMs benefit from their strong multimodal understanding and instruction-following capabilities, enabling them to adapt to diverse and complex task requirements. Therefore, an increasing body of literature \cite{jiang2024vlm2vec, meng2025vlm2vec, gu2025breaking, yu2025cafe, thirukovalluru2025breaking} proves that MLLMs can be used to learn Universal Multimodal Embedding (UME) that captures general-purpose content similarity. Meanwhile, evaluation benchmarks such as the Multimodal Embedding Benchmark (MMEB) \cite{jiang2024vlm2vec} and its upgraded version MMEB-V2 \cite{meng2025vlm2vec} have addressed this academic demand for UME research, covering 78 instruction-aware tasks across three modalities.

\begin{figure}[htbp]
    \centering
    \includegraphics[width=\linewidth]{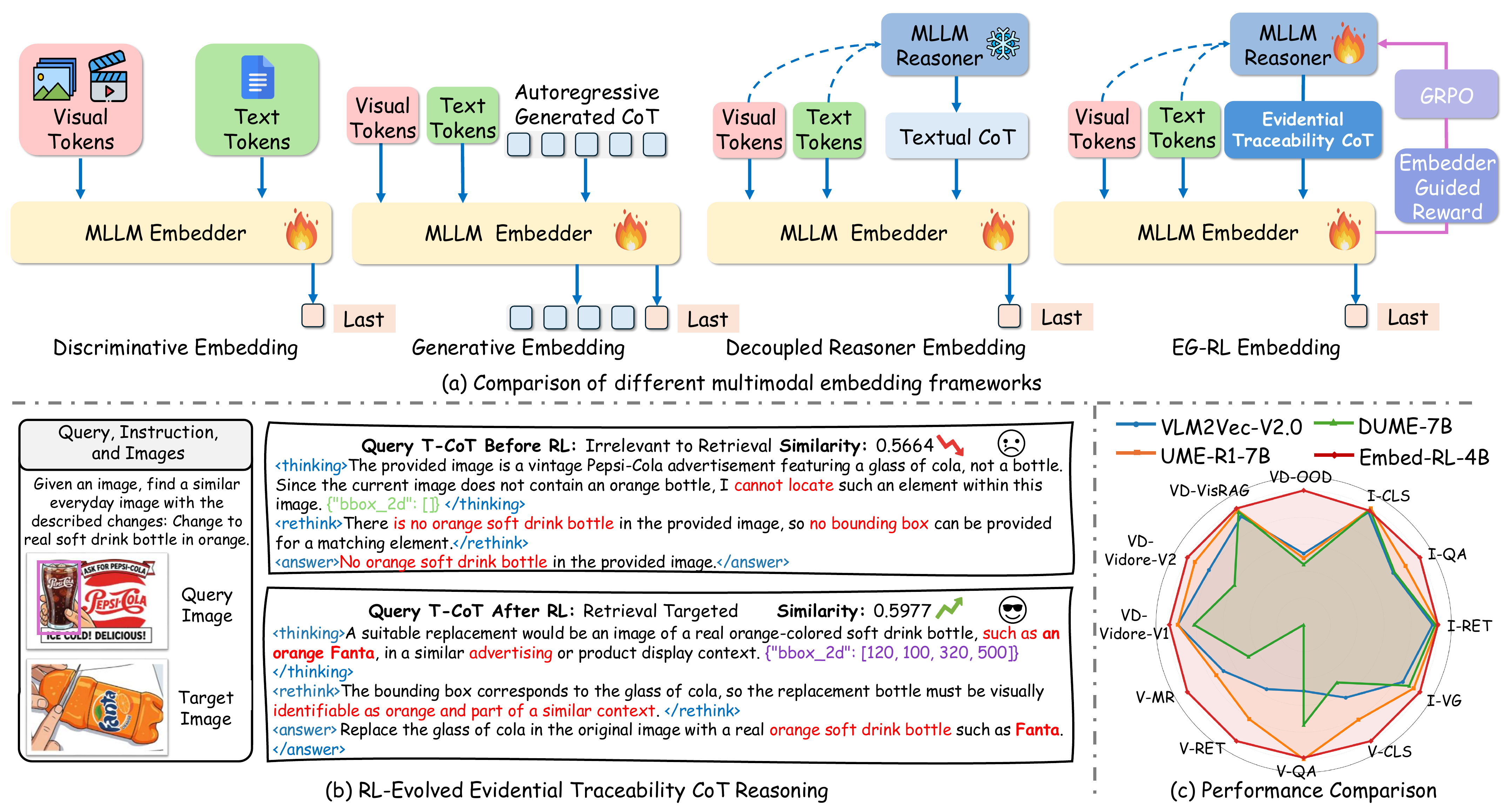}
    \caption{Multimodal embedding optimization via Embedder-Guided Reinforcement Learning (EG-RL). (a) Frameworks evolution. (b) Reasoning enhancement with RL-optimized evidential Traceability CoT (T-CoT). (c) Comparison of multi-task performance.}
    \label{fig:teaser}
\end{figure}

Currently, most MLLM-powered embedding methods are discriminative embedding models \cite{liu2025lamra, zhang2024gme, lan2025llave}. These models typically extract embedding features directly from the final hidden states of input tokens, failing to fully leverage MLLMs’ inherent generative capabilities and reasoning potential. Consequently, recent studies explore integrating generative reasoning into UME tasks, as shown in Figure \ref{fig:teaser}. For instance, approaches like UME-R1 \cite{lan2025ume} unify discriminative and generative embeddings through textual Chain-of-Thoughts (CoTs) generated by the MLLM Embedder. However, simultaneously optimizing contrastive loss and next-token prediction objectives causes conflicting gradients, leading to suboptimal performance \cite{chen2025moca, lan2025ume}. In contrast, the decoupled Reasoner-Embedder paradigm proposed by TTE \cite{cui2025think}, which relies on massive computational resources and data distillation from large models, aims to alleviate this problem by decoupling the two processes, using pre-trained MLLMs to generate offline CoT reasoning to enhance embedding quality with only the Embedder trained. Nevertheless, the CoTs generated by the Reasoner of TTE are not specifically designed for embedding, as they are not trained together with the Embedder. This misalignment may introduce noise and even lead to hallucinations. Moreover, relying only on textual reasoning fails to fully leverage the MLLM Embedder’s potential to process multimodal signals, whose rich representations could significantly enhance retrieval performance. Such insufficiency of multimodal cues leads to notable embedding alignment bias, where critical visual-spatial cues and video-temporal signals are not fully captured in retrieval, resulting in less effective cross-modal matching and restricted generalization on complex real-world multimodal tasks.

To address the above issues, this paper proposes a reasoning-driven decoupled UME framework. This framework leverages the Embedder-Guided Reinforcement Learning (EG-RL) algorithm to optimize the CoTs generated by the Reasoner, using our novel process reward oriented to the alignment between query and target and verifiable outcome reward for retrieval. Firstly, we construct a dataset that initially trains the Embedder to generate high-quality embeddings conditioned on the sequence of preceding input and CoT tokens. The trained Embedder acts as a reward model and provides stable and reliable reward signals. Secondly, inspired by the region-aware paradigm \cite{su2025thinking, zheng2025deepeyes, fan2025grit, wang2025traceable} that makes the model focus on the region of interest (RoI), we propose the evidential Traceability CoT (T-CoT) that explicitly guides the Embedder to focus on task-related information, effectively filter out redundant visual elements, and integrate modality-specific critical cues to adapt to long-text retrieval, coarse-grained semantic matching, and fine-grained alignment for robust performance across heterogeneous tasks. The main contributions of this paper are summarized as follows:
\begin{enumerate}

    \item \textbf{Embedder-Guided Reinforcement Learning}. We propose a novel decoupled RL framework where the Embedder guides the Reasoner to optimize CoT trajectories for specific embedding tasks. This approach resolves key conflict between generative and embedding objectives, ensures the Reasoner’s output greatly improves retrieval quality, and addresses core challenge of adapting general CoTs to embedding tasks.
    
    \item \textbf{Evidential Traceability CoT for Embedding}. We further extend CoT reasoning to complex multimodal scenarios, integrating explicit visual localization information, video keyframes, and text keywords into detailed inference trajectories. This design enables the model to focus on core retrieval-related information and effectively mitigate the negative impact of redundant multimodal and text data on overall embedding alignment performance.
    
    \item \textbf{Efficient Performance Improvement Across Multiple Benchmarks}. Under computationally constrained settings, the framework proposed in this paper outperforms state-of-the-art generative embedding models on both MMEB-V2 \cite{meng2025vlm2vec} and video retrieval UVRB \cite{guo2025towards} benchmark datasets, and achieves exceptional performance across diverse combinatorial scenarios.

\end{enumerate}
\section{Related Work}

\subsection{Universal Multimodal Embedding}

Constructing robust multimodal representations is a fundamental core challenge in multimodal learning. Pioneering models such as CLIP \cite{radford2021learning} and ALIGN \cite{jia2021scaling} adopt a dual-encoder architecture and learn effective representations through contrastive learning on rich large-scale image-text paired data. However, they struggle to handle interleaved image-text inputs, and their text encoders lack sufficient capacity to understand truly complex textual content. To address this issue, researchers leverage Multimodal Large Language Models (MLLMs) to build embedding models \cite{jiang2024vlm2vec, lin2024mm, meng2025vlm2vec, liu2025lamra, zhou2025megapairs, zhang2024gme}, capitalizing on their strong multimodal comprehension capabilities to enhance overall learning performance.

Existing works focus on different aspects: VLM2Vec \cite{jiang2024vlm2vec} transforms MLLMs into embedding models via contrastive learning and achieves outstanding performance on unconventional retrieval tasks such as visual question answering and localization; MM-Embed \cite{lin2024mm} explores using off-the-shelf MLLMs as zero-shot rerankers to optimize retrieval results; LamRA \cite{liu2025lamra} unifies the multimodal retrieval paradigm through two-stage retrieval training and joint reranking. MegaPairs \cite{zhou2025megapairs} and GME \cite{zhang2025bridging} address the modality imbalance problem with automated pipelines; LLaVE \cite{lan2025llave} and Unite \cite{kong2025modality} focus on hard negative sample mining.

Recent studies focus on instruction-aware representations: MMEB \cite{jiang2024vlm2vec} and MMEB-V2 \cite{meng2025vlm2vec} construct a comprehensive evaluation benchmark covering 78 tasks. UME-R1 \cite{lan2025ume} first introduces reasoning mechanisms, yet simultaneous optimization of dual components via Reinforcement Learning (RL) leads to conflicts, and redundant Chain-of-Thought (CoT) trajectories dilute representations. TTE \cite{cui2025think} adopts a decoupled yet computationally expensive architecture, and its Reasoner is misaligned with retrieval tasks, resulting in task-irrelevant outputs. In this paper, we propose a decoupled RL framework enabling separate optimization of dual components and generating retrieval-relevant reasoning trajectories through a dual reward mechanism, addressing the aforementioned challenges.

\subsection{Multimodal Reasoning with Reinforcement Learning}

MLLMs \cite{liu2023visual, li2024llava, wang2024qwen2, Qwen3-VL} extend and enrich the capabilities of Large Language Models (LLMs) to the multimodal domain, achieving promising results across diverse tasks including visual question answering \cite{liu2023visual, bai2025qwen2, wang2025svqa, geng2025webwatcher}, visual grounding \cite{lai2024lisa, chung2025don, li2025dyfo}, and keyframe extraction \cite{zhang2025thinking, wang2025vr, zhu2025vau, liao2025improved}. Early works \cite{xu2025llava, wei2022chain, zhang2024chain, wang2022self, ji2024tree} mostly completed reasoning tasks using standardized CoT prompts. Since DeepSeek-R1 \cite{guo2025deepseek} proposed the Group Relative Policy Optimization‌ (GRPO) RL algorithm, numerous recent works have optimized advanced RL algorithms \cite{yu2025dapo, zheng2025group} and enhanced the reasoning capabilities of MLLMs \cite{wu2025vtool, cao2025ground, duan2025got, su2025openthinkimg, wang2025vg}.

GRIT \cite{fan2025grit} interleaves bounding box coordinates with textual reasoning chains and designs an RL scheme based on the GRPO algorithm, enabling efficient training with dual robust rewards and no additional annotated data. Ground-R1 \cite{cao2025ground} proposes an RL framework to achieve grounded visual reasoning without extra annotations, guiding response generation through dual rewards to improve reasoning reliability and interpretability. BRPO \cite{chu2025qwen} uses Intersection over Union (IoU)-based rewards to guide models to autonomously generate visual-text reflections, combined with a visual token mechanism to mitigate the problems of visual attention dilution and hallucinations. DeepEyes \cite{zheng2025deepeyes} adopts end-to-end RL to induce models to develop the ability of ``thinking with images'', improving performance on various reasoning tasks. TreeVGR \cite{wang2025traceable} proposes the TreeBench benchmark and the TreeVGR training paradigm, enhancing visual grounding reasoning capabilities by jointly supervising localization and reasoning via RL. Inspired by grounding reasoning, this paper further proposes evidential Traceability CoT (T-CoT), which constructs structured multimodal reasoning chains by extracting bounding boxes of images, keyframes of videos, and keywords of text. This method enables the model to focus on the core regions of retrieval tasks, thereby improving embedding quality.

\section{Methodology}
\label{sec:methodology}

\subsection{Preliminaries}
\label{sec:preliminaries}
We focus on the universal multimodal retrieval task, where given a query $q$ (text, image, or interleaved text-image modalities) and a candidate set $\Omega = \{c_n\}_{n=1}^N$, the goal is to retrieve the most relevant candidate from $\Omega$.

\begin{figure}[ht!]
    \centering
    \includegraphics[width=\linewidth]{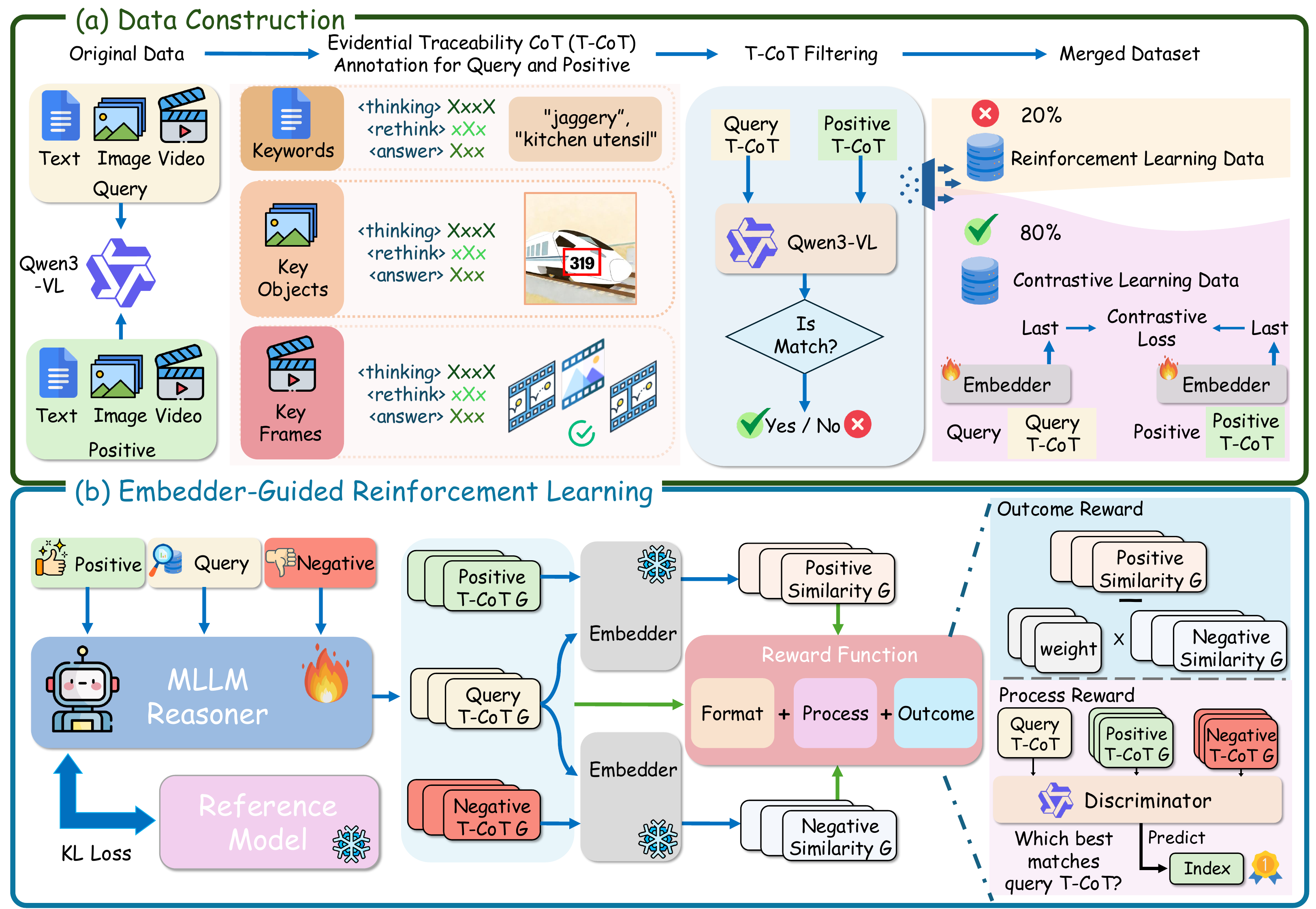}
    \caption{Overview of the proposed data synthesis and EG-RL framework. (a) Data Construction generates T-CoT annotations for query-positive pairs, filters and splits the dataset to enable contrastive and reinforcement learning, laying the groundwork for reasoning-aware embedding. (b) Embedder-Guided Reinforcement Learning finetunes the MLLM with a process-outcome reward function, encouraging T-CoT trajectories that yield more discriminative and beneficial generative embeddings.}
    \label{fig:pipeline}
\end{figure}

To learn discriminative multimodal embeddings, we adopt contrastive learning with the InfoNCE loss \cite{oord2018representation}, as shown in Figure~\ref{fig:pipeline}(a). For a query $q_i$, its positive target $t_i^+$, and its negative target set $\mathcal{T}^- = \{t_j^-\}_{j \neq i}$, the InfoNCE loss optimizes the model to maximize the similarity between $q_i$ and $t_i^+$ while minimizing the similarities to all $t_j^- \in \mathcal{T}^-$. The loss is defined as:

\begin{equation}
\scalebox{1}{
$\mathcal{L}_{\text{InfoNCE}} = -\frac{1}{N} \sum_{i=1}^N \log \frac{\exp\left(\cos(\boldsymbol{h}_{q_i}, \boldsymbol{h}_{t_i^+}) / \tau\right)}{\exp\left(\cos(\boldsymbol{h}_{q_i}, \boldsymbol{h}_{t_i^+}) / \tau\right) + \sum_{t^- \in \mathcal{T}^-} \exp\left(\cos(\boldsymbol{h}_{q_i}, \boldsymbol{h}_{t^-}) / \tau\right)}$
},
\end{equation}
where $\boldsymbol{h}_{q_i}$ and $\boldsymbol{h}_{t}$ are embeddings of $q_i$ and target $t$ (extracted as the last-layer hidden states of the last token from a vision-language model), $\cos(\cdot, \cdot)$ denotes cosine similarity, and $\tau$ is the temperature hyperparameter.

\subsection{Data Construction}
\label{sec:data_construction}
To support the training of reasoning-driven universal multimodal embeddings, we construct a high-quality multimodal dataset following a ``sampling-annotation-filtering-splitting'' pipeline, as illustrated in Figure~\ref{fig:pipeline}(a). The dataset integrates diverse modalities (image, video, visual document) and ensures alignment between reasoning trajectories and retrieval objectives through strict quality control.

We first curate the initial data pool by adopting a stratified sampling strategy across three core sources, and referencing the data paradigm of VLM2Vec-V2~\cite{meng2025vlm2vec}: (1) Image-centric tasks from MMEB-train~\cite{jiang2024vlm2vec}, covering image classification, Question Answering, retrieval, and grounding; (2) Video-language instruction data from LLaVA-Hound~\cite{zhang2025direct}, including video captioning, QA, and retrieval; (3) Visual document retrieval data from ViDoRe~\cite{faysse2024colpali} and VisRAG~\cite{yu2024visrag}.

Next, we perform evidential Traceability Chain-of-Thought (T-CoT) annotation for all query-positive pairs. Each T-CoT follows a structured three-part format: (1) \texttt{<thinking>} extracts modality-specific cues (text keywords via \texttt{text\_keywords}, image spatial locations via \texttt{bbox\_2d} (two-dimensional bounding box), video critical moments via \texttt{key\_frames}); (2) \texttt{<rethink>} refines reasoning logic to focus on key retrieval-relevant aspects; (3) the final answer summarizes core retrieval-relevant information. We design task-specific prompts for annotation, ensuring T-CoT aligns with diverse multimodal retrieval scenarios.

Following annotation, we perform a strict CoT-guided relevance filtering to eliminate noisy samples. With a custom-designed judgment prompt, we assess whether the T-CoTs of queries and their positive samples are clearly irrelevant or contradictory to the task description. We retain only samples labeled ``No'', meaning those that are relevant and not contradictory, for contrastive learning. This filtering step effectively mitigates noise interference in contrastive learning. The initial dataset contains 2.22 million samples, and 1.83 million are preserved after filtering, yielding a retention rate of approximately 80\%. Approximately 20\% of the filtered-out samples are uniformly sampled and used in the reinforcement learning stage, as these hard examples are valuable for model exploration in reinforcement learning. In addition, we assign training weights to different datasets based on their task importance and data quality.

The constructed dataset exhibits three features: (1) modal diversity, covering three modalities of text, image, and video; (2) reasoning alignment, where T-CoT explicitly integrates multimodal cues and retrieval-related logic, avoiding informational redundancy; (3) quality assurance, as rigorous filtering and weighted sampling ensure that the dataset is free of significant noise and balanced across tasks. This dataset lays a solid foundation for the two-stage training, enabling Embedder to learn reasoning-aware representations and the Reasoner to optimize the generation of retrieval-centric T-CoTs through reinforcement learning.

\subsection{Embedder-Guided Reinforcement Learning}
\label{sec:embed_guided_rl}
To address the misalignment between generative reasoning and embedding objectives, we propose a decoupled reinforcement learning framework in which the pre-trained Embedder provides supervision to the Reasoner. This framework optimizes the generation of T-CoT to prioritize retrieval-relevant multimodal cues, leveraging a dual-guidance reward mechanism and Group Relative Policy Optimization (GRPO)~\cite{guo2025deepseek}. Figure \ref{fig:pipeline}(b) illustrates the workflow of this stage.

\textbf{EG-RL Framework Design.}
First, we fully train an Embedder using the InfoNCE loss to equip it with robust embedding capabilities.
Our RL framework maintains strict separation between two components: the Reasoner, which is responsible for generating T-CoT, and the Embedder, which is frozen after contrastive training. This decoupling ensures three key benefits: (1) targeted optimization of reasoning without disrupting the Embedder's learned discriminative capabilities; (2) stable reward signals from the frozen Embedder that consistently evaluate T-CoT quality based on embedding alignment; and (3) flexible integration of multi-source rewards to internalize both retrieval and reranking knowledge. The Reasoner takes multimodal queries as input and outputs structured T-CoT, which integrates three critical cues: text keywords, image bounding boxes, and video keyframes. Additionally, we recrop the content within the bounding boxes and keyframes based on T-CoT to achieve multimodal reasoning-aware embeddings. This structured reasoning is then concatenated with the original input to form the Embedder's input, denoted as $\mathcal{I}$:
\begin{equation}
\mathcal{I} = \left[ x_{\text{text}}, \, x_{\text{img}}, \, x_{\text{vid}}, \, \text{T-CoT}(x), \, \text{\textless emb\textgreater} \right].
\end{equation}
In this equation, $\text{\textless emb\textgreater}$ is a special token whose hidden state is extracted as the final embedding, and the evaluation of this embedding by the Embedder directly guides the Reasoner's policy update.

\textbf{Reward Function with Process and Outcome Guidance.}
We design a three-component reward function to align T-CoT generation with embedding quality, combining format compliance, outcome-level retrieval effectiveness, and process-level T-CoT alignment:

\textbf{\emph{Format Reward ($\mathcal{R}_{\text{format}}$)}}: This reward ensures T-CoT strictly follows the predefined template (\texttt{<thinking>} $\rightarrow$ \texttt{<rethink>} $\rightarrow$ \texttt{<answer>}) and includes all required multimodal cues. Reward 1 for full compliance, 0 otherwise, guaranteeing T-CoT output interpretability and compatibility with the Embedder module.

\textbf{\emph{Embedder-Guided Outcome Reward ($\mathcal{R}_{\text{outcome}}$)}}: 
This reward measures how T-CoT improves embedding alignment by jointly assessing the ranking accuracy of positive samples and the similarity margin between positive and hard negative samples. The margin is a softmax-weighted average of negative similarities scaled by a temperature parameter. For a query $q_i$ with positive target $t_i^+$ and in-batch negatives $\{t_j^-\}_{j \neq i}$, embeddings are $e_{q_i} = \pi_e(q_i, o_i^q)$ and $e_{t_j} = \pi_e(t_j, o_j^t)$, where $o_i^q$ and $o_j^t$ are T-CoT outputs for query and target. The reward for $o_i^q$ is defined as:
\begin{equation}
\mathcal{R}_{\text{outcome}}(o_i^q) = \operatorname{Acc}_{k}(e_{q_i}, t_i^+) \cdot \Big( \text{sim}(e_q, e_{t_i^+}) - \mathbb{E}_{\tau}\big[\text{sim}(e_{q_i}, e_{t_j^-})\big] \Big),
\end{equation}
where $\operatorname{Acc}_{k}(e_{q_i}, t_i^+)$ denotes the top-$k$ retrieval accuracy, which measures whether $t_i^+$ is among the top-$k$ ranked targets when sorted by cosine similarity to $e_{q_i}$; $\text{sim}(\cdot, \cdot)$ denotes the cosine similarity between normalized embeddings; and $\mathbb{E}_{\tau}[\cdot]$ stands for the softmax-weighted average of cosine similarities between $e_{q_i}$ and embeddings of in-batch negative targets.

Additionally, we compute $\mathcal{R}_{\text{outcome}}$ symmetrically for positive targets: taking $t_i^+$ as the anchor, $q_i$ as its positive query, and other in-batch queries as negatives to calculate $\mathcal{R}_{\text{outcome}}$ for $o_i^t$. This symmetric computation enforces consistent embedding alignment in both query-to-target and target-to-query directions. This reward optimizes T-CoT with embedding learning as the core objective, enhancing its discriminative ability across samples.

\textbf{\emph{T-CoT Process Reward ($\mathcal{R}_{\text{process}}$)}}:
We employ an independent pretrained Vision-Language Model Discriminator $\mathcal{D}$ for listwise comparison to align T-CoT outputs of queries and targets. Let $q_{\text{cot}}$ be the query's T-CoT output and $\{c_{\text{cot}}^j\}_{j=1}^m$ the T-CoT outputs of $m$ in-batch candidate targets, comprising positive samples from multiple rollouts of the query's data pair and negative samples. After shuffling, the index set of ground-truth positives is denoted $\mathcal{P}$. To mitigate position bias, we feed $q_{\text{cot}}$ and shuffled $\{c_{\text{cot}}^j\}_{j=1}^m$ to $\mathcal{D}$ as pairwise inputs. The reward quantifies alignment via $\mathcal{D}$'s selection correctness, formally defined as:
\begin{equation}
\mathcal{R}_{\text{process}}(o_i) =
\begin{cases}
1, & \text{if } \mathcal{D}\big(q_{\text{cot}}, \{c_{\text{cot}}^j\}_{j=1}^m\big) \in \mathcal{P}, \\
0, & \text{otherwise},
\end{cases}
\end{equation}
where $o_i$ denotes the T-CoT generation outcome of the $i$-th sample, and $\mathcal{D}(\cdot, \cdot)$ outputs the index of the candidate T-CoT most aligned with $q_{\text{cot}}$ in the shuffled candidate set.
A reward of 1 indicates that $\mathcal{D}$ correctly selects a positive T-CoT from the ground-truth set, signifying well-aligned query-target T-CoT pairs; a reward of 0 means $\mathcal{D}$ fails to select any positive T-CoT, indicating misalignment between query and target T-CoT outputs.
We compute $\mathcal{R}_{\text{process}}$ symmetrically in the reverse direction, moving from positive targets to their corresponding queries, to ensure consistent embedding alignment across both directions.

This reward guides Reasoner to align query and target T-CoT outputs.
Since T-CoT is the intermediate process for embedding generation, T-CoT alignment quantified by $\mathcal{D}$ selection correctness directly improves embedding quality.

The total reward is a weighted combination of these three components:
\begin{equation}
\mathcal{R}_{\text{total}} = \alpha\mathcal{R}_{\text{format}} + \beta\mathcal{R}_{\text{process}} + \gamma\mathcal{R}_{\text{outcome}, }
\end{equation}
where $\alpha, \beta, \gamma $ are non-negative weighting coefficients that balance the relative contributions of the three reward components in the total reward optimization.

\textbf{Policy Optimization with GRPO.}
We adopt GRPO to optimize the Reasoner’s policy, and use group-based rewards to stabilize the training process. For each query-target pair $q \sim \mathcal{S}$, where $\mathcal{S}$ denotes the training sample set of query-target pairs. We sample $G=8$ candidate T-CoT sequences $\{o_i\}_{i=1}^G$ according to the old policy $\pi_{\theta_{\text{old}}}$. The optimization objective is defined as:

\begin{equation}
\begin{split}
\mathcal{L}_{\text{grpo}} &= 
\mathbb{E}_{\substack{q \sim \mathcal{S}, \\ \{o_i\}\sim\pi_{\theta_{\text{old}}}}} 
\Bigg[ 
\frac{1}{G} \sum_{i=1}^G \Big( 
    \quad \min\mspace{-3mu}\left(r_{\theta}(o_i)A_i, \text{clip}\mspace{-2mu}\left(r_{\theta}(o_i), 1-\epsilon, 1+\epsilon\right)A_i\right) \\
&\quad \quad \quad \quad \quad \quad \quad \quad \quad \quad - \beta \mathbb{D}_{\text{KL}}\mspace{-3mu}\left(\pi_{\theta} \| \pi_{\text{ref}}\right) 
\Big)
\Bigg],
\end{split}
\end{equation}
where $r_{\theta}(o_i) = \pi_{\theta}(o_i|q)/\pi_{\theta_{\text{old}}}(o_i|q)$ denotes importance ratio, $\epsilon$ is the clipping threshold of importance ratio, $\beta$ is a hyperparameter weighting the Kullback-Leibler divergence term, $\pi_{\text{ref}}$ denotes the reference policy model before optimization, and $A_i = (r_i - \mu_r)/\sigma_r$ represents advantage, where $\mu_r = \text{mean}(\{r_1,...,r_G\})$ and $\sigma_r = \text{std}(\{r_1,...,r_G\})$ are the mean and std of group rewards.
\section{Experiments}
\label{sec:experiments}

\subsection{Implementation Details}
We train Qwen3-VL-2B~\cite{Qwen3-VL} and Qwen3-VL-4B~\cite{Qwen3-VL} as Embedders with the DeepSpeed Zero2 optimization framework, and adopt a sub-batch strategy following VLM2Vec~\cite{jiang2024vlm2vec}. The models are trained for 2 epochs with a learning rate of 1e-4 and weight decay of 0.01; the batch size is computationally light: 512 for the 2B model and 256 for the 4B model, and we use Low-Rank Adaptation (LoRA)~\cite{hu2022lora} for fine-tuning. For reinforcement learning, Qwen3-VL-8B~\cite{Qwen3-VL} is trained as Reasoner via the GRPO algorithm~\cite{guo2025deepseek} for 1 epoch, with a computationally light batch size of 256, a learning rate of 3e-6, and standard GRPO hyperparameters.

\subsection{Baselines and Datasets}
We compare with representative multimodal embedding models with diverse architectures, modalities and scales.
These baselines cover image, video and visual document retrieval, ensuring thorough and fair evaluation.
We evaluate against GME~\cite{zhang2025bridging}, ColPali~\cite{faysse2024colpali}, VLM2Vec~\cite{jiang2024vlm2vec}, LamRA~\cite{liu2025lamra}, CAFe~\cite{yu2025cafe}, VLM2Vec-V2~\cite{meng2025vlm2vec}, TTE~\cite{cui2025think},UME-R1~\cite{lan2025ume}, InternVideo2~\cite{wang2024internvideo2}, Unite~\cite{gu2025breaking}, and GVE~\cite{guo2025towards}.

For the training phase, we followed VLM2Vec-V2~\cite{meng2025vlm2vec} and constructed a comprehensive dataset from three key sources: video-language instruction data from LLaVA-Hound~\cite{zhang2025direct}, visual document retrieval data from ViDoRe~\cite{faysse2024colpali} and VisRAG~\cite{yu2024visrag}, and image-based vision-language task data from MMEB-train~\cite{jiang2024vlm2vec}. Detailed training procedures, hyperparameters, and dataset construction are in the supplementary material. We evaluate on two comprehensive benchmarks:

\textbf{MMEB-V2} (Massive Multimodal Embedding Benchmark)\cite{meng2025vlm2vec}: It is a comprehensive and robust benchmark consisting of 78 diverse tasks across three core visual modalities (image, video, and visual document). MMEB-V2 extends the original MMEB \cite{jiang2024vlm2vec} by introducing five additional meta-tasks focused specifically on video and visual document understanding, bringing the total to nine meta-tasks. We adopted Hit@1 as the evaluation metric for image and video tasks, and Normalized Discounted Cumulative Gain (NDCG@5)~\cite{jarvelin2002cumulated} for visual documents.

\textbf{UVRB} (Universal Video Retrieval Benchmark) \cite{guo2025towards}: It is a suite of 16 datasets to identify capability gaps in video retrieval across tasks and domains. UVRB measures multi-dimensional generalization over textual, composite, and visual retrieval tasks, including across coarse-grained, fine-grained, and long-context scenarios. We report mean Average Precision (mAP) for all UVRB tasks.

\begin{table}[htbp]
\caption{Comparison of performance between baselines and our method on MMEB-V2. CLS: classification, QA: question answering, RET: retrieval, GD: grounding, MRET: moment retrieval, VDR: ViDoRe, VR: VisRAG, OOD: out-of-domain. The highest and second-highest values are highlighted in bold and underline.}
\vspace{2mm}
\centering
\renewcommand{\arraystretch}{1.2}
\resizebox{\textwidth}{!}{
\begin{tabular}{l ccccc ccccc ccccc c}
\toprule
\multirow{2}{*}{Model} & \multicolumn{5}{c}{Image} & \multicolumn{5}{c}{Video} & \multicolumn{5}{c}{VisDoc} & \multirow{2}{*}{All} \\
\cmidrule(lr){2-6} \cmidrule(lr){7-11} \cmidrule(lr){12-16}
& CLS & QA & RET & GD & Overall & CLS & QA & RET & MRET & Overall & VDRv1 & VDRv2 & VR & OOD & Overall \\
\midrule
\# of Datasets & 10 & 10 & 12 & 4 & 36 & 5 & 5 & 5 & 3 & 18 & 10 & 4 & 6 & 4 & 24 & 78 \\
\midrule
\rowcolor[HTML]{EDEDED}
\multicolumn{17}{c}{\emph{Baseline Models}} \\
\midrule
ColPali-V1.3-3B~\cite{faysse2024colpali} & 40.3 & 11.5 & 48.1 & 40.3 & 34.9 & 26.7 & 37.8 & 21.6 & 25.5 & 28.2 & 83.6 & 52.0 & 81.1 & 43.1 & 71.0 & 44.4 \\ 
GME-2B~\cite{zhang2025bridging} & 54.4 & 29.9 & 66.9 & 55.5 & 51.9 & 34.9 & 42.0 & 25.6 & 32.4 & 33.9 & \underline{86.1} & \underline{54.0} & 82.5 & 43.1 & 72.7 & 54.1 \\ 
GME-7B~\cite{zhang2025bridging} & 57.7 & 34.7 & 71.2 & 59.3 & 56.0 & 37.4 & 50.4 & 28.4 & 38.2 & 38.6 & \textbf{89.4} & \textbf{55.6} & \textbf{85.0} & 44.4 & \textbf{75.2} & 57.8 \\ 
LamRA-2-7B~\cite{liu2025lamra} & 59.2 & 26.5 & 70.0 & 62.7 & 54.1 & 39.3 & 42.6 & 24.3 & 34.6 & 35.2 & 22.0 & 11.5 & 37.4 & 21.0 & 23.9 & 40.4 \\ 
LamRA-2.5-7B~\cite{liu2025lamra} & 51.7 & 34.1 & 66.9 & 56.7 & 52.4 & 32.9 & 42.6 & 23.2 & 37.6 & 33.7 & 56.3 & 33.3 & 58.2 & 40.1 & 50.2 & 47.4 \\ 
VLM2Vec-2B~\cite{jiang2024vlm2vec} & 58.7 & 49.3 & 65.0 & 72.9 & 59.7 & 33.4 & 30.5 & 20.6 & 33.0 & 29.0 & 49.8 & 13.5 & 51.8 & 33.5 & 41.6 & 47.0 \\ 
VLM2Vec-7B~\cite{jiang2024vlm2vec} & 62.7 & 56.9 & 69.4 & 82.2 & 65.5 & 39.1 & 30.0 & 29.0 & 40.6 & 34.0 & 56.9 & 9.4 & 59.1 & 38.1 & 46.4 & 52.3 \\ 
VLM2Vec-V2-2B~\cite{meng2025vlm2vec} & 62.9 & 56.3 & 69.5 & 77.3 & 64.9 & 39.3 & 34.3 & 28.8 & 38.5 & 34.9 & 75.5 & 44.9 & 79.4 & 39.4 & 65.4 & 58.0 \\ 
VLM2Vec-V2-7B~\cite{meng2025vlm2vec} & 65.7 & 61.5 & 70.0 & 85.2 & 68.1 & 45.9 & 33.9 & 27.6 & 39.3 & 36.4 & 78.8 & 52.6 & 82.7 & 42.1 & 69.3 & 61.2 \\ 
CAFe-7B~\cite{yu2025cafe} & 63.6 & 61.7 & 69.1 & 87.6 & 67.6 & 35.8 & \underline{58.7} & 34.4 & 39.5 & 42.4 & 70.7 & 49.6 & 79.5 & 38.1 & 63.9 & 60.6 \\ 
$\text{TTE}_s$-2B~\cite{cui2025think} & \textbf{67.9} & 66.6 & 70.2 & 84.1 & \underline{70.1} & 47.3 & 49.1 & 34.4 & 33.2 & 32.1 & 77.5 & 53.2 & 83.2 & 41.1 & 68.8 & 63.1 \\ 
UME-R1-2B~\cite{lan2025ume} & 64.8 & 62.8 & 67.6 & 77.2 & 66.6 & 44.3 & 51.2 & 32.9 & 39.7 & 42.2 & 72.4 & 46.2 & 79.2 & 37.2 & 63.9 & 60.1 \\ 
UME-R1-7B~\cite{lan2025ume} & \underline{67.1} & \underline{69.2} & \textbf{71.9} & 84.9 & \textbf{71.3} & 48.6 & \textbf{60.7} & \underline{38.2} & 39.3 & 47.5 & 75.7 & 50.5 & 83.7 & 37.6 & 67.1 & 64.5 \\ 

\midrule
\rowcolor[HTML]{FFE5E5}
\multicolumn{17}{c}{\emph{Ours}} \\
\midrule
Embed-RL-2B & 62.8 & 67.9 & 68.6 & \underline{90.4} & 69.2 & \underline{57.0} & 55.9 & \textbf{45.1} & \underline{49.4} & \underline{52.1} & 79.9 & 52.0 & 84.6 & \underline{65.7} & 74.1 & \underline{66.8} \\ 
Embed-RL-4B & 63.7 & \textbf{70.5} & \underline{71.3} & \textbf{91.4} & \underline{70.1} & \textbf{57.6} & 58.4 & \textbf{45.1} & \textbf{49.5} & \textbf{53.0} & 80.2 & 53.4 & \underline{84.9} & \textbf{67.1} & \underline{74.7} & \textbf{68.1} \\ 

\bottomrule
\end{tabular}
}
\label{tab:main_result}
\end{table}

\subsection{Main Results}

\begin{wraptable}[14]{r}{0.4\textwidth}
\vspace{-33pt}
\centering
\footnotesize
\caption{Video retrieval performance on UVRB. Domain dimensions: Coarse-grained (CG), Fine-grained (FG), Long-context (LC).
The best and second-best results are marked in \textbf{bold} and \underline{underline}.}
\label{tab:selected_domains}
\resizebox{0.4\textwidth}{!}{
\begin{tabular}{l|ccc}
\toprule
\textbf{Model} & \textbf{CG} & \textbf{FG} & \textbf{LC} \\
\midrule
InternVideo2-6B~\cite{wang2024internvideo2}  & 50.4 & 41.7 & 42.3 \\
VLM2Vec-V2~\cite{meng2025vlm2vec}       & 49.8 & 50.2 & 76.2 \\
GME-7B~\cite{zhang2025bridging}           & 51.8 & 50.7 & 78.8 \\
Unite-7B~\cite{kong2025modality}         & 54.1 & 53.9 & 74.6 \\
GVE-3B~\cite{guo2025towards}           & 55.2 & 54.1 & 76.4 \\
\midrule
Embed-RL-2B      & \underline{59.1} & \underline{54.6} & \textbf{86.9} \\
Embed-RL-4B      & \textbf{60.7} & \textbf{55.6} & \underline{86.1} \\
\bottomrule
\end{tabular}
}
\end{wraptable}

Table~\ref{tab:main_result} compares the comprehensive performance of our proposed Embed-RL models with various baseline approaches on the MMEB-V2 benchmark. Our Embed-RL models consistently achieve superior performance directly compared to all baseline models. Specifically, Embed-RL-4B attains the best overall score of 68.1, outperforming the strong next top baseline UME-R1-7B by 3.6 points. Embed-RL-2B follows closely with an overall score of 66.8, also surpassing all baseline variants. Across different modalities, our models show clear and notable advantages. In the Image modality, Embed-RL-4B impressively achieves the best grounding (GD) performance of 91.4, with Embed-RL-2B ranking second at 90.4. For the Video modality, both Embed-RL-2B and Embed-RL-4B outperform all baselines in overall score, with Embed-RL-2B scoring 52.1 and Embed-RL-4B scoring 53.0, and they achieve a video retrieval (RET) score of 45.1. In the visual document modality, we observe significant improvements in out-of-domain (OOD) performance, where Embed-RL-4B reaches 67.1 and Embed-RL-2B 65.7, far exceeding prior baseline results. These results demonstrate the effectiveness of our proposed approach across diverse visual modalities and task types.

In the broader field of video retrieval, benefiting from the effectiveness of our T-CoT to accurately locate keywords or keyframes, our model exhibits significant advantages in Coarse-grained (CG), Fine-grained (FG), and Long-context (LC) retrieval tasks. As shown in Table \ref{tab:selected_domains}, detailing video retrieval performance of different models on the UVRB dataset across the three domains, our Embed-RL models consistently excel: 4B tops CG at 60.7 and FG at 55.6, while 2B leads LC at 86.9, both outperforming all existing baselines by a clear margin.

\begin{figure}[htbp]
    \centering
    \includegraphics[width=\linewidth]{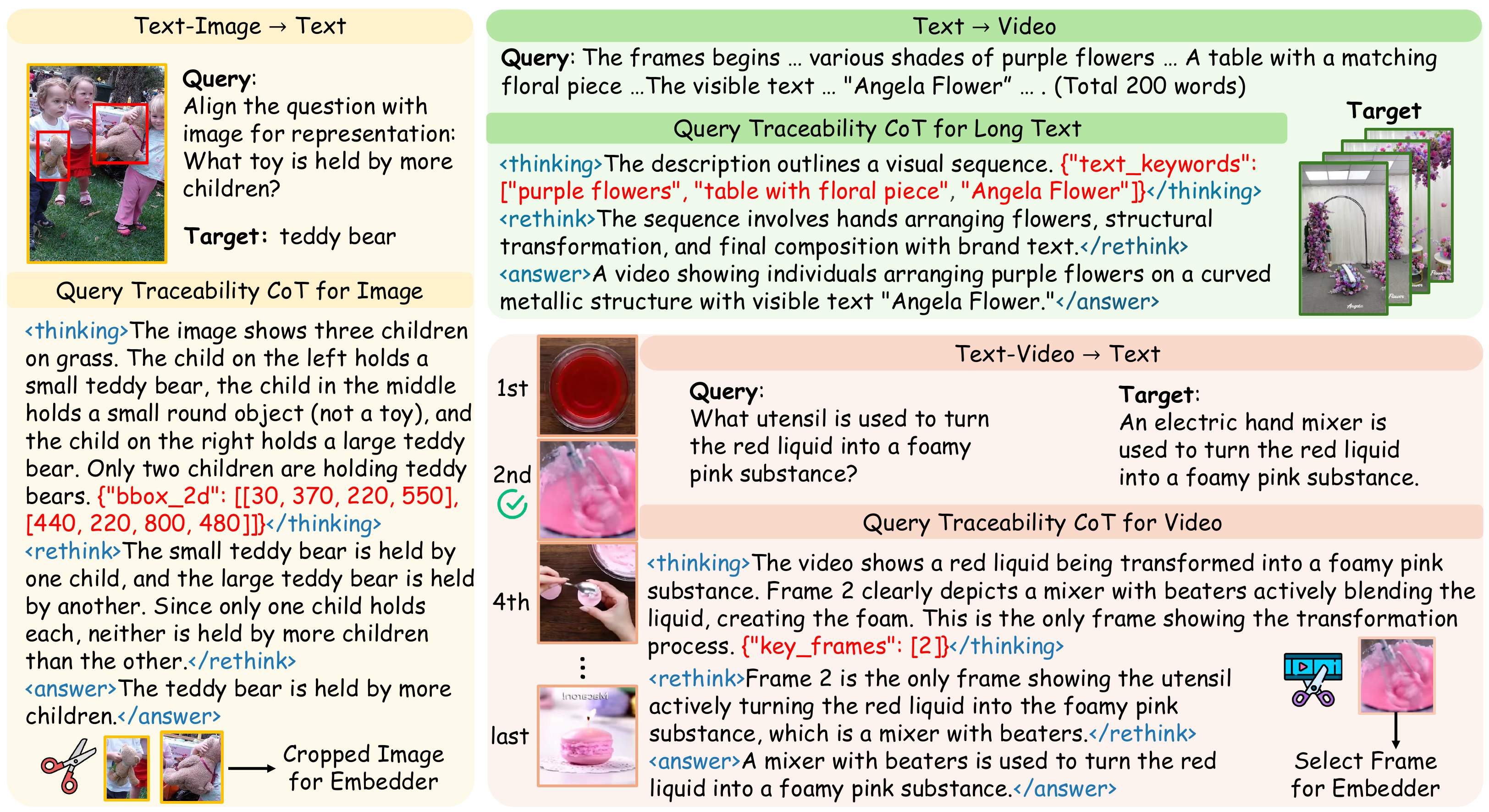}
    \caption{Example visualization of our reasoning-driven embedding framework on multimodal retrieval tasks. The figure shows the evidential Traceability CoT reasoning process for video and visual document retrieval.}
    \label{fig:example}
\end{figure}

Additionally, Figure~\ref{fig:example} presents the visualization of our T-CoT on text, image and video; we crop bbox and keyframe to achieve multi-modal CoT input, and our T-CoT accurately locates retrieval needs to improve retrieval performance. More scores and visualizations are provided in the supplementary material.

\subsection{Ablation Study}
To dissect the contribution of each component in our framework, we conduct ablation experiments on MMEB-V2, where Embed-RL-2B serves as the model.

\textbf{Analysis of the reward model in the RL Stage.}As shown in Table~\ref{tab:ablation_rewards_1}, removing the RL stage leads to an overall performance decline of 1.5 points, dropping the score from 66.8 to 65.3. This result verifies that RL fine-tuning is indispensable for embedding alignment. Omitting weighted negative sampling, a core module of our contrastive reward mechanism, brings a performance reduction of 0.3 points from 66.8 to 66.5. This outcome emphasizes the component’s key function in prioritizing hard negative examples to strengthen discriminative embedding learning.The process reward is formulated to reward logical reasoning steps and align query T-CoT with target T-CoT. It contributes 0.8 points to the overall performance, as its exclusion lowers the score from 66.8 to 66.0. This component shows the most notable influence on video tasks, where performance falls from 52.1 to 51.3. This trend reveals that video understanding strongly depends on step-by-step reasoning and further reflects the critical role of the process reward in T-CoT alignment.Additionally, the outcome reward is built to reward final predictions. It contributes 1.0 point to the overall performance, and its removal reduces the total score from 66.8 to 65.8. This reward ensures that the reasoning process remains consistent with the objective of the target task.

\textbf{Impact of Reasoning Components on T-CoT.} As shown in Table~\ref{tab:ablation_rewards_2}, removing the reasoning process while retaining the answer part leads to a performance decrease of 1.3 points from 66.8 to 65.5. Image grounding and video moment retrieval see notable drops, falling from 69.2 to 67.9 and 52.1 to 50.5, which highlights the importance of multimodal evidence tracking for fine-grained alignment tasks. Removing multimodal cues leads to a performance reduction of 1.0 points from 66.8 to 65.8, validating the necessity of extracting multimodal cues through bounding boxes and keyframes to enhance alignment between multimodal representations and retrieval objectives. Most critically, using only raw input without T-CoT causes a catastrophic decrease in overall performance of 6.6 points, from 66.8 to 60.2. The greatest impact appears on video tasks, where performance falls from 52.1 to 43.7. This dramatic decline demonstrates the necessity of high-quality evidential Traceability CoT for retrieval accuracy, as it enables the model to decompose complex retrieval and understanding tasks into manageable steps and improve cross-modal embedding quality.

\begin{table}[htbp]
\begin{minipage}[t]{0.49\textwidth}
\makeatletter\def\@captype{table}
\centering
\tabcolsep=3pt
\scriptsize
\caption{Ablation Study on Reward Components in EG-RL stage.}
\label{tab:ablation_rewards_1}
\begin{adjustbox}{width=\textwidth}
\begin{tabular}{p{2.95cm}<{\centering}p{0.7cm}<{\centering}p{0.7cm}<{\centering}p{0.7cm}<{\centering}p{0.7cm}<{\centering}}
\toprule
\textbf{Model} & \textbf{Image} & \textbf{Video} & \textbf{VisDoc} & \textbf{All} \\
\midrule
Embed-RL-2B & \textbf{69.2} & \textbf{52.1} & \textbf{74.1} & \textbf{66.8} \\
\midrule
w/o EG-RL & 68.0 & 50.1 & 72.7 & 65.3 \\
w/o weighted negative & 68.9 & 51.7 & 73.9 & 66.5 \\
w/o process reward & 68.3 & 51.3 & 73.5 & 66.0 \\
w/o outcome reward & 68.1 & 51.2 & 73.1 & 65.8 \\
\bottomrule
\end{tabular}
\end{adjustbox}
\end{minipage}
\hspace{0.01\textwidth}
\begin{minipage}[t]{0.48\textwidth}
\makeatletter\def\@captype{table}
\centering
\tabcolsep=3pt
\scriptsize
\caption{Ablation Study on Reasoning Components in T-CoT.}
\label{tab:ablation_rewards_2}
\begin{adjustbox}{width=\textwidth}
\begin{tabular}{p{2.6cm}<{\centering}p{0.7cm}<{\centering}p{0.7cm}<{\centering}p{0.7cm}<{\centering}p{0.7cm}<{\centering}}
\toprule
\textbf{Model} & \textbf{Image} & \textbf{Video} & \textbf{VisDoc} & \textbf{All} \\
\midrule
Embed-RL-2B & \textbf{69.2} & \textbf{52.1} & \textbf{74.1} & \textbf{66.8} \\
\midrule
w/o reasoning & 67.9 & 50.5 & 73.1 & 65.5 \\
w/o multimodal cues & 68.1 & 51.4 & 73.3 & 65.8 \\
w/ raw input & 60.4 & 43.7 & 72.4 & 60.2 \\
\bottomrule
\end{tabular}
\end{adjustbox}
\end{minipage}
\end{table}

\begin{figure}[htbp]
    \centering
    \includegraphics[width=\linewidth]{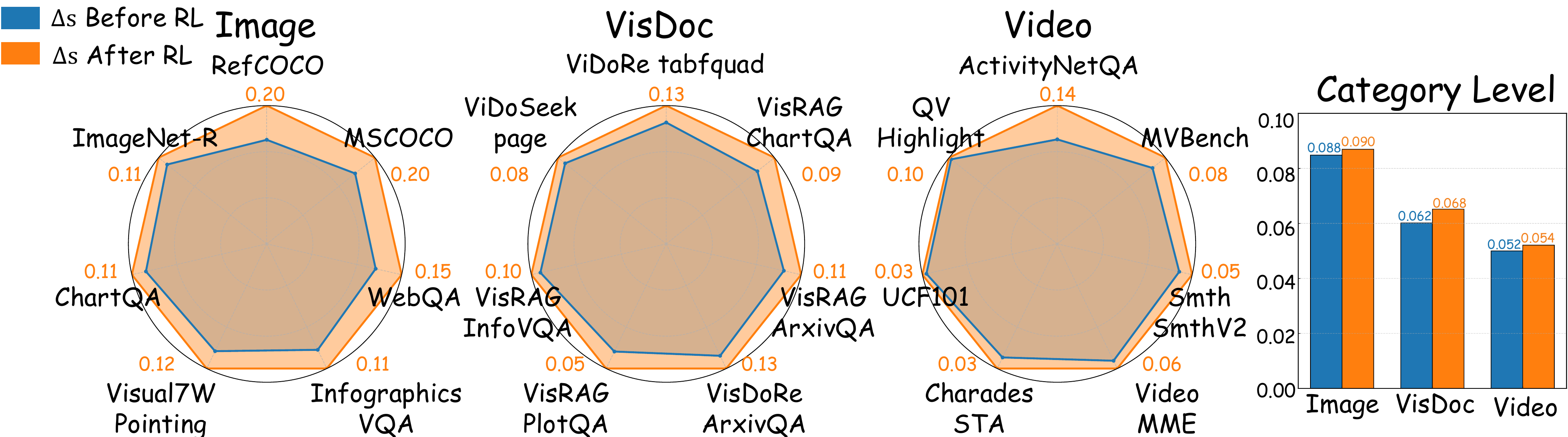}
\caption{Similarity difference \( \Delta s = \text{sim}(\text{query}, \text{top1}) - \text{sim}(\text{query}, \text{top2}) \) before and after EG-RL.
Here, \( \text{sim}(\cdot,\cdot) \) denotes cosine similarity of normalized embeddings,
\(\text{top1}\) is the most similar positive candidate and \(\text{top2}\) the second-most similar.
This metric quantifies the model’s discriminative ability over similar candidates on multimodal datasets.}
    \label{fig:diff}
\end{figure}

\textbf{Ablation Study on Model’s Discriminative Ability for Candidates.}
We define the top-ranked candidates with the highest similarity (excluding positive samples) as highly similar candidate samples.
On this basis, we study how optimizing the reasoner with EG-RL improves the model’s ability to distinguish between similar candidates.
Specifically, we calculate the difference between the similarity of the query to the most similar candidate and that to the second-most similar candidate on each dataset, both before and after RL. This difference measures whether the model assigns a significantly higher similarity to positive samples than to other highly similar candidates.
As shown in Figure~\ref{fig:diff}, we observe that on different datasets across three modalities, the radar chart obtained after RL prominently encloses the one obtained before RL. This indicates that the computed similarity difference becomes larger after RL, widening the gap between the query’s similarity to the top-ranked candidate and the second-ranked candidate. It demonstrates that the model’s ability to discriminate between similar candidates is effectively enhanced.
Meanwhile, the bar chart shows that the model achieves consistent overall improvement on three-modality datasets. This verifies that optimizing T-CoT with RL strengthens the model’s general ability to distinguish between different candidates.

\textbf{Ablation on traceable evidence count and retrieval metrics.}We also systematically analyze the relationship between the number of traceable evidence pieces and core retrieval metrics across all datasets before and after reinforcement learning.For images and visual document data, we count the change in the number of bounding boxes.For video data, we similarly count the change in the number of keyframes.We observe that after reinforcement learning, the T-CoT generated by the Reasoner tends to produce more bounding boxes.For the video modality, the model tends to focus on fewer keyframes.The retrieval metrics yield consistent and significant improvements, with the curves after reinforcement learning lying entirely above those before.For the image modality, the model captures more visual evidence to boost reasoning accuracy and recall.For the video modality, it concentrates on critical frames and conducts precise keyframe extraction and temporal localization to identify key content.These changes are particularly pronounced on complex samples involving multi-object localization and multi-person relationship reasoning.

\begin{figure}[htbp]
    \centering
    \includegraphics[width=\linewidth]{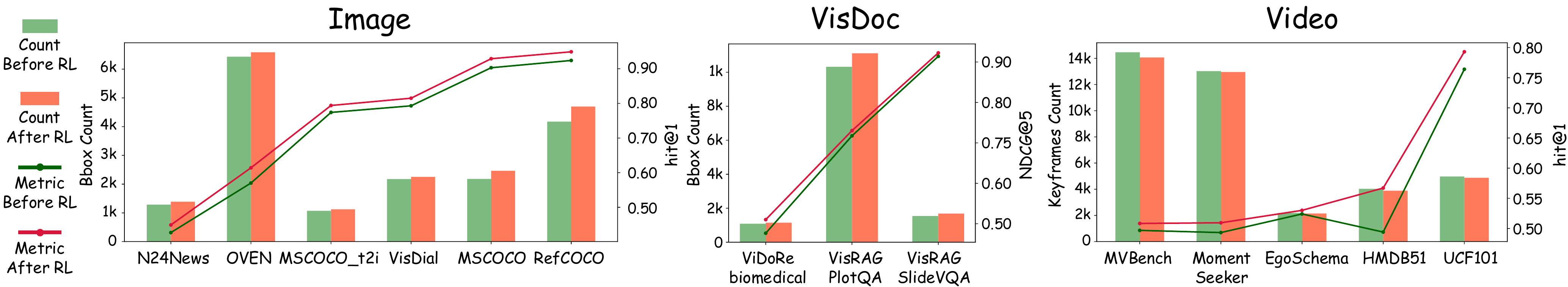}
    \caption{Relationship between traceable evidence counts and retrieval metrics across datasets.
    Hit@1 is employed for Image and Video;
    NDCG@5 is used for VisDoc.
    Bounding box counts are shown for Image and VisDoc, while keyframe counts for Video.}
    \label{fig:count}
\end{figure}

\section{Conclusion}
This work addresses key limitations of generative universal multimodal embedding (UME) methods: chain-of-thought (CoT) remains text-only, resulting in poor retrieval relevance, while joint optimization of generative and embedding objectives gives rise to gradient conflicts that impede cross-modal matching. 
We propose Embed-RL, a reasoning-driven UME model built upon Embedder-Guided RL (EG-RL), which serves as a decoupled reinforcement learning framework that integrates multimodal evidential Traceability CoT (T-CoT) and a retrieval-oriented dual-reward mechanism to enable precise reasoning-embedding alignment.
Extensive experiments on the MMEB-V2 and UVRB benchmarks demonstrate that Embed-RL outperforms state-of-the-art counterparts within computational constraints, achieving significant improvements in cross-modal retrieval and out-of-domain generalization tasks. This work shows that targeted reasoning optimization can substantially enhance multimodal embeddings, providing an efficient solution for reasoning-driven UME and valuable insights into the integration of generative reasoning with multimodal representation learning.

\clearpage

\begin{figure}[htbp]
    \centering
    \includegraphics[width=\linewidth]{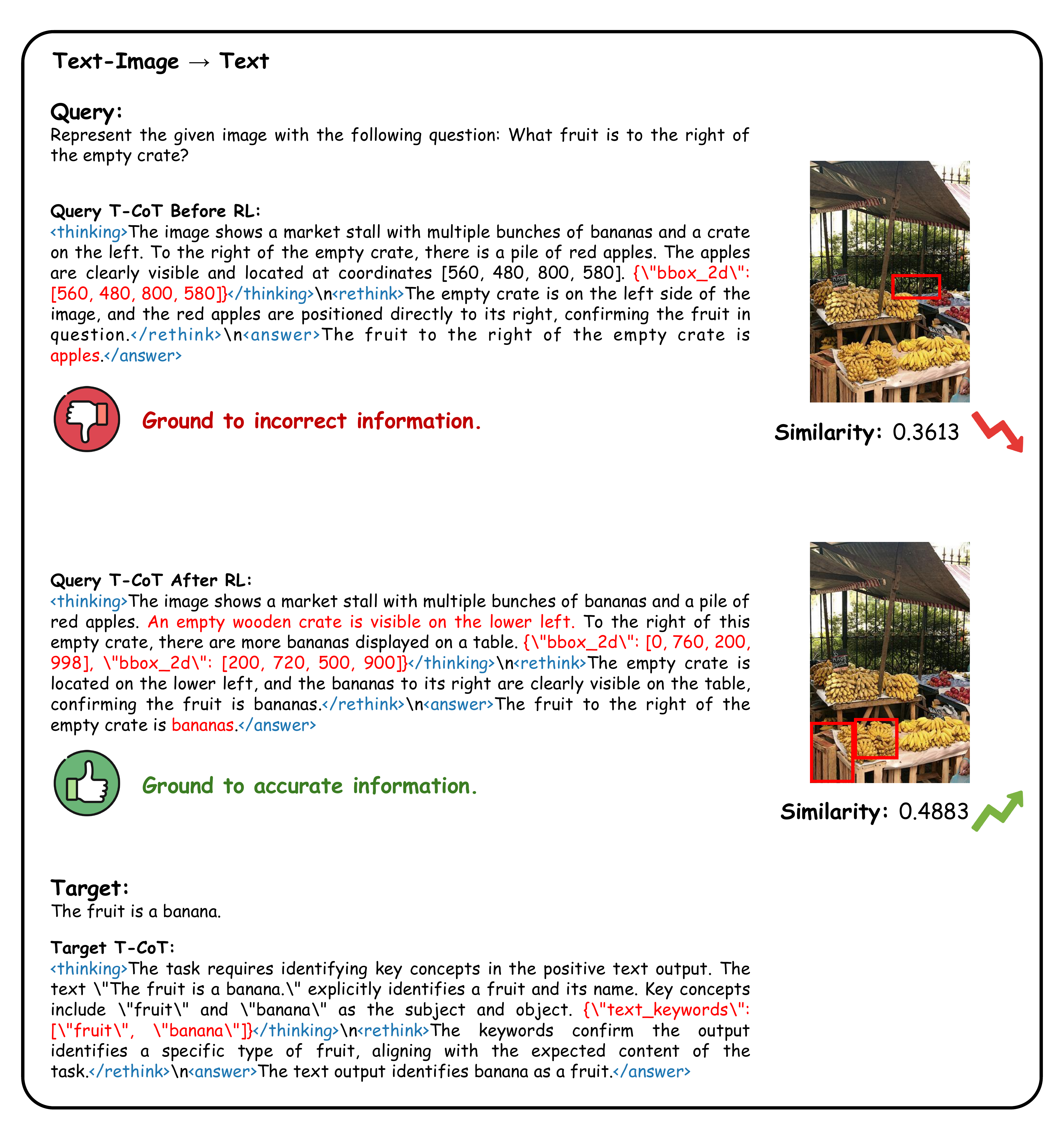}
    \caption{Example 1 of T-CoT Before and After EG-RL.}
    \label{fig:sup_example_2}
\end{figure}

\clearpage

\begin{figure}[htbp]
    \centering
    \includegraphics[width=\linewidth]{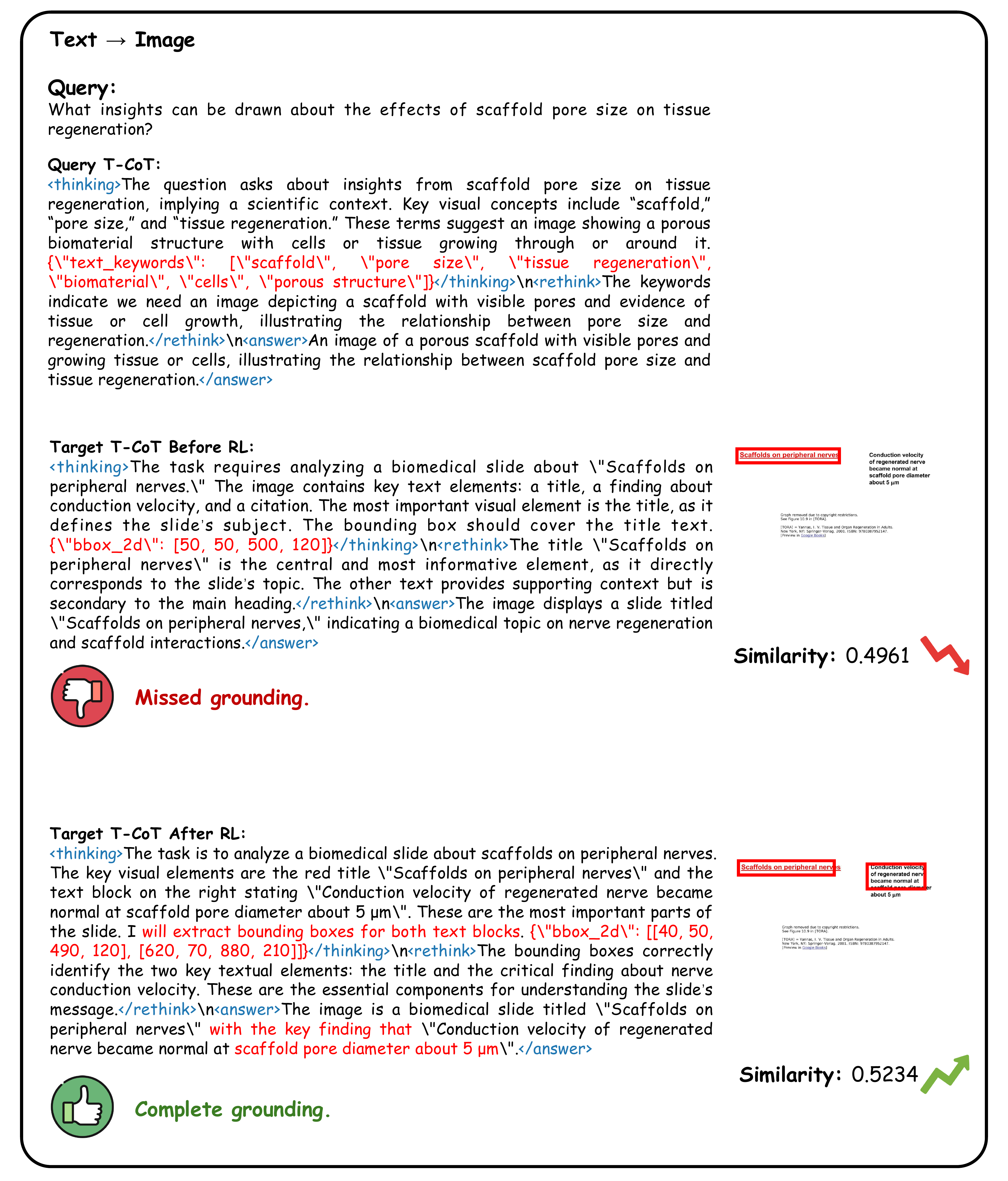}
    \caption{Example 2 of T-CoT Before and After EG-RL.}
    \label{fig:sup_example_4}
\end{figure}

\clearpage

\begin{figure}[htbp]
    \centering
    \includegraphics[width=\linewidth]{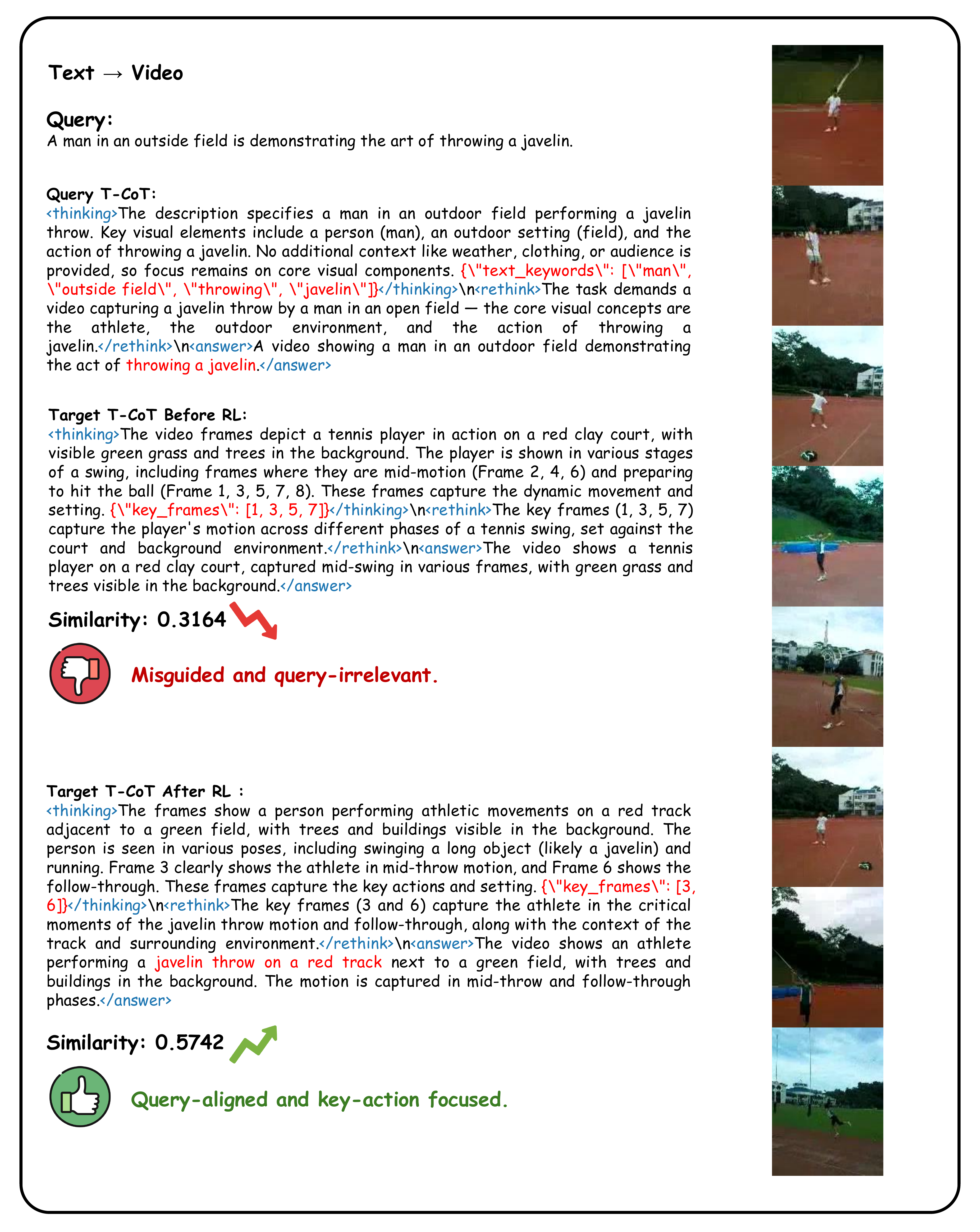}
    \caption{Example 3 of T-CoT Before and After EG-RL.}
    \label{fig:sup_example_7}
\end{figure}

\clearpage

\bibliographystyle{splncs04}
\bibliography{main}

\clearpage

\appendix
\definecolor{prompt-bg}{RGB}{245,250,245}
\definecolor{prompt-border}{RGB}{60,120,100}
\definecolor{prompt-title-bg}{RGB}{60,120,100}
\definecolor{prompt-title-fg}{RGB}{255,255,255}
\definecolor{prompt-leftbar}{RGB}{80,140,120}

\lstdefinestyle{promptstyle}{
    basicstyle=\sffamily\scriptsize,
    breaklines=true,
    backgroundcolor=\color{prompt-bg},
    frame=none, numbers=none,
    columns=fullflexible,
    keepspaces=true,
    showspaces=false,
    showstringspaces=false,
    tabsize=2,
    escapeinside=||,
    lineskip=-0.5pt,
}

\newtcolorbox{promptbox}[1]{
    enhanced, breakable,
    arc=2pt, boxrule=0.6pt,
    colframe=prompt-leftbar!50!prompt-border,
    colback=prompt-bg,
    coltitle=prompt-title-fg,
    fonttitle=\bfseries\scriptsize,
    title={\strut #1},
    titlerule=0pt,
    toptitle=2pt, bottomtitle=1pt,
    top=3pt, bottom=3pt, left=6pt, right=5pt,
    boxsep=0pt,
    colbacktitle=prompt-title-bg,
    before title={\faCode\hspace{0.25em}},
}

{
\centering
\Large
\textbf{Embed-RL: Reinforcement Learning for Reasoning-Driven Multimodal Embeddings} \\
\vspace{0.5em}Supplementary Material \\
\vspace{1.0em}
}

\section{Additional details}
\label{app:additional-details}
\setcounter{subsection}{-1}

In this supplementary material, we elaborate on further insights, provide detailed derivations, and include additional qualitative results to enhance the comprehensiveness of our work.

\section{Training Details}
\label{app:training_details}
\subsection{Contrastive Learning}
We trained the Qwen3-VL-2B-Instruct~\cite{Qwen3-VL} and Qwen3-VL-4B-Instruct~\cite{Qwen3-VL} models using the DeepSpeed Zero2 optimization framework. Key training hyperparameters were summarized as follows:
The training process was conducted for 2 epochs with a batch size of 512 for Qwen3-VL-2B-Instruct and 256 for Qwen3-VL-4B-Instruct. Following the sub-batch training scheme in VLM2Vec~\cite{jiang2024vlm2vec}, we adopted a sub-batch training strategy that ensured samples in each sub-batch are drawn from the same dataset,
where the sub-batch size was set to 256 for Qwen3-VL-2B-Instruct and 128 for Qwen3-VL-4B-Instruct. We set the initial learning rate to 1e-4, using a cosine learning rate scheduler with 10 warm-up steps and a weight decay of 0.01. We employed Low-Rank Adaptation (LoRA)~\cite{hu2022lora} for fine-tuning:
the rank $r$ and scaling factor $\alpha$ were set to 64 and 128 for Qwen3-VL-2B-Instruct, and 96 and 192 for Qwen3-VL-4B-Instruct, respectively. For comparison, UME-R1~\cite{lan2025ume} was trained with a batch size of 1024 and TTE~\cite{cui2025think} was trained with 8192, indicating that our training scale was significantly smaller than both methods. Notably, our experimental results could be further scaled up with more abundant computational resources.

\subsection{Embedder-Guided Reinforcement Learning}

In the reinforcement learning stage, we trained the Qwen3-VL-8B-Instruct~\cite{Qwen3-VL} Reasoner with the GRPO~\cite{guo2025deepseek}. We adopted in-batch negative contrastive rewards for optimization, with GRPO hyperparameters set as group size $G = 8$, clipping parameter $\varepsilon = 0.2$, and KL-divergence coefficient $\beta = 0.01$. We set batch size to 256, learning rate to $3\text{e-}6$, and trained the model for one epoch. We also restricted each step to samples from the same dataset to avoid overly simple negative samples affecting optimization.

For the embedder-guided outcome reward $\mathcal{R}_{\text{outcome}}$, two core hyperparameters balance retrieval accuracy and similarity margin calculation: the top-$k$ parameter for retrieval accuracy $\operatorname{Acc}_k$ and the temperature parameter $\tau$ for softmax-weighted negative sampling.

Top-$k$ Retrieval Accuracy $k$:
We set the parameter $k$ in $\operatorname{Acc}_k(e_{q_i}, t_i^+)$ to 8. Specifically, $\operatorname{Acc}_8(e_{q_i}, t_i^+)$ checks whether all 8 T-CoT rollouts of a query-target pair rank in the top-8 by cosine similarity to $e_{q_i}$, which captures the embedding alignment consistency critical for stable GRPO reward optimization.

Temperature Parameter ($\tau$):
The temperature $\tau$ scales the softmax weights for hard negative sampling in the similarity margin calculation, which is defined as:
\begin{equation}
\mathbb{E}_{\tau}\big[\text{sim}(e_{q_i}, e_{t_j^-})\big] = \frac{\sum_{j \neq i} \exp\left(\frac{\text{sim}(e_{q_i}, e_{t_j^-})}{\tau}\right) \cdot \text{sim}(e_{q_i}, e_{t_j^-})}{\sum_{j \neq i} \exp\left(\frac{\text{sim}(e_{q_i}, e_{t_j^-})}{\tau}\right)}.
\end{equation}
We set $\tau = 0.5$ for all experiments, a value chosen to emphasize hard negatives while avoiding overfitting to noisy negative samples.

For process reward, we employ an independent Qwen3-VL-8B-Instruct~\cite{Qwen3-VL} as the pretrained VLM Discriminator $\mathcal{D}$. It performs listwise comparison to align query T-CoT outputs with corresponding target outputs. The discriminator selects the candidate T-CoT that best matches the query. Selection correctness quantifies alignment quality and forms the process reward signal.

We set $\alpha=0.05$, $\beta=0.8$, and $\gamma=0.2$ for the total reward $\mathcal{R}_{\text{total}}$ to balance the contributions of format, process, and outcome rewards respectively.

\subsection{Multimodel Vision Processing}
For visual input processing, we set specific pixel constraints for images and videos to balance computational efficiency and feature fidelity:

For images: MIN\_PIXELS = $128 \times 32 \times 32$ and MAX\_PIXELS = $768 \times 32 \times 32$.

For videos: VIDEO\_MIN\_PIXELS = $128 \times 32 \times 32$, VIDEO\_MAX\_PIXELS = $300 \times 32 \times 32$, and VIDEO\_TOTAL\_PIXELS = $300 \times 32 \times 32 \times 8$.

Video frame sampling hyperparameters were fixed as FRAME\_FACTOR = 2, FPS = 2.0, FPS\_MIN\_FRAMES = 8 and FPS\_MAX\_FRAMES = 8.

For multimodal cropping, we first convert the relative coordinates of bounding boxes, which are scaled to the range 0–1000, into the original image coordinates, then conduct cropping on the raw image. Keyframes corresponding to the sampled frames are re-extracted and concatenated between the \textless /thinking\textgreater and \textless rethink\textgreater tokens.

\clearpage
\section{Detailed Dataset Construction}
\label{app:data_construct}

\subsection{Data Sources and Initial Sampling Strategy}
To enable effective multi-modal task training, we adopt the training data paradigm of VLM2Vec-V2~\cite{meng2025vlm2vec} and build a comprehensive dataset from three core sources: (1) video-language instruction data (LLaVA-Hound~\cite{zhang2025direct}), (2) visual document retrieval data (ViDoRe~\cite{faysse2024colpali} and VisRAG~\cite{yu2024visrag}), and (3) image-based vision task data (MMEB-train~\cite{jiang2024vlm2vec}). We apply a stratified sampling strategy across data modalities to ensure balanced coverage:

\begin{itemize}
\item Image-based datasets: Maximum 50,000 samples per dataset
\item Document-based datasets: Maximum 100,000 samples per dataset
\item Video-based datasets: Maximum 300,000 samples per dataset
\end{itemize}

Full sampling is used if the original dataset size is smaller than the above maximum. As shown in Table~\ref{tab:cot_filter_stats}, we presented the exact number of samples selected for each dataset in our experiments.

\begin{table*}[htbp]
    \centering
    \setlength{\tabcolsep}{2pt}
    \small
    \caption{Statistics of Initial Sampling and CoT-guided Filtering.}
    \label{tab:cot_filter_stats}
    \resizebox{\linewidth}{!}{
        \begin{tabular}{@{}cccccc@{}}
            \toprule
            \textbf{Dataset} & \textbf{Initial Samples} & \textbf{Filtered Samples} & \textbf{Retention Ratio} & \textbf{Weight} & \textbf{Modality} \\
            \midrule
            
            \rowcolor[HTML]{EDEDED}
            \multicolumn{6}{c}{\textit{Image-based (MMEB-train)}} \\
            A-OKVQA & 50,000 & 37,929 & 75.86\% & 0.26 & Text-Image → Text \\
            CIRR & 50,000 & 35,085 & 70.17\% & 0.43 & Text-Image → Text-Image \\
            ChartQA & 50,000 & 39,512 & 79.02\% & 0.35 & Text-Image → Text \\
            DocVQA & 50,000 & 47,401 & 94.80\% & 0.84 & Text-Image → Text \\
            \textcolor{gray}{HatefulMemes} & \textcolor{gray}{25,500} & \textcolor{gray}{16,572} & \textcolor{gray}{64.99\%} & \textcolor{gray}{0.30} & \textcolor{gray}{Text-Image → Text} \\
            ImageNet-1K & 50,000 & 44,409 & 88.82\% & 2.25 & Text-Image → Text \\
            InfographicsVQA & 50,000 & 40,746 & 81.49\% & 0.31 & Text-Image → Text \\
            MSCOCO & 50,000 & 26,429 & 52.86\% & 3.78 & Text-Image → Text-Image \\
            MSCOCO-i2t & 50,000 & 46,596 & 93.19\% & 2.58 & Text-Image → Text \\
            MSCOCO-t2i & 50,000 & 43,173 & 86.35\% & 2.32 & Text → Text-Image \\
            \textcolor{gray}{N24News} & \textcolor{gray}{50,000} & \textcolor{gray}{30,320} & \textcolor{gray}{60.64\%} & \textcolor{gray}{1.65} & \textcolor{gray}{Text-Image → Text} \\
            NIGHTS & 47,823 & 43,167 & 90.26\% & 0.23 & Text-Image → Text-Image \\
            OK-VQA & 27,027 & 19,900 & 73.63\% & 0.25 & Text-Image → Text \\
            SUN397 & 50,000 & 45,864 & 91.73\% & 0.22 & Text-Image → Text \\
            \textcolor{gray}{VOC2007} & \textcolor{gray}{23,532} & \textcolor{gray}{20,454} & \textcolor{gray}{86.92\%} & \textcolor{gray}{0.24} & \textcolor{gray}{Text-Image → Text} \\
            Visual7W & 50,000 & 41,677 & 83.35\% & 1.68 & Text-Image → Text \\
            VisDial & 50,000 & 34,652 & 69.30\% & 3.75 & Text → Text-Image \\
            VisualNews-i2t & 50,000 & 34,364 & 68.73\% & 2.91 & Text-Image → Text \\
            VisualNews-t2i & 50,000 & 28,684 & 57.37\% & 3.49 & Text → Text-Image \\
            WebQA & 50,000 & 43,910 & 87.82\% & 0.23 & Text → Text-Image \\
            
            \rowcolor[HTML]{EDEDED}
            \multicolumn{6}{c}{\textit{Video-based (LLaVA-Hound)}} \\
            Caption Retrieval & 300,000 & 283,721 & 94.57\% & 5.27 & Video → Text \\
            Video QA & 300,000 & 273,906 & 91.30\% & 4.38 & Video-Text → Text \\
            Video Retrieval & 300,000 & 260,410 & 86.80\% & 5.76 & Text → Video \\
            
            \rowcolor[HTML]{EDEDED}
            \multicolumn{6}{c}{\textit{Document-based}} \\
            ViDoRe & 100,000 & 83,964 & 83.96\% & 5.0 & Text-Image → Text \\
            VisRAG & 100,000 & 60,266 & 60.27\% & 6.0 & Text → Image \\
            
            \midrule
            \textbf{Image-based} & 1,123,882 & 865,074 & 76.97\% & - & Image-centric \\
            \textbf{Video-based} & 900,000 & 819,037 & 91.02\% & - & Video-centric \\
            \textbf{Document-based} & 200,000 & 144,230 & 72.12\% & - & Document-centric \\
            \textbf{Total} & 2,223,882 & 1,828,341 & 82.21\% & - & Multimodal \\
            \bottomrule
        \end{tabular}
    }
\end{table*}

We exclude three classification datasets (HatefulMemes, N24News, VOC2007) from the first-stage contrastive learning. This stage only employs contrastive loss under a single-dataset sub-batch constraint. Their limited number of categories results in false negatives and noise with large sub-batches, which impairs representation learning. This explains the model’s weak performance on image classification, as it never encounters these samples during training.

\subsection{CoT-guided Relevance Filtering}We generate chain-of-thought (CoT) annotations for queries and positive samples using Qwen3-VL-8B~\cite{Qwen3-VL}, and further conduct strict relevance filtering via a custom prompt with the same model to discard annotations that are irrelevant to or conflict with the query task. To mitigate noise in contrastive learning, only samples labeled "No" are retained. Table~\ref{tab:cot_filter_stats} lists the sample size, retention ratio, training weight, and modality of each dataset.

\begin{promptbox}{ CoT Relevance and Conflict Judgment Prompt}
\begin{lstlisting}[style=promptstyle]
Your task is to determine whether the content of `query_cot` and `pos_cot` are **obviously irrelevant** or **obviously conflicting** based on the given `qry` (task description).
**Rules**:
1. Only output a single word: "Yes" (if obviously irrelevant/conflicting) or "No" (if relevant and not conflicting)
2. "Obviously irrelevant": The content of query_cot and pos_cot have no logical connection to each other or to the qry task
3. "Obviously conflicting": The core conclusions/key information of query_cot and pos_cot are mutually contradictory
4. Only judge "obvious" cases - if the relevance is ambiguous, output "No"
\end{lstlisting}
\end{promptbox}

\subsection{RL Dataset Sampling}
Using the high-quality CoT-filtered dataset, we apply equidistant sampling to construct the reinforcement learning (RL) training set, ensuring uniform distribution across challenging sub-datasets, as shown in Table~\ref{tab:rl_sampling_stats}.

\begin{table}[htbp]
    \centering
    \setlength{\tabcolsep}{4pt}
    \small
    \caption{Reinforcement Learning Dataset Construction and Sampling Strategy.}
    \label{tab:rl_sampling_stats}
    \resizebox{0.7\linewidth}{!}{
    \begin{tabular}{lll}
        \toprule
        \textbf{Sampled Dataset} & \textbf{RL Samples} & \textbf{Modality} \\
        \midrule
        A-OKVQA & 1,000 & Text-Image → Text \\
        llavahound video retrieval & 2,000 & Text → Video \\
        ViDoRe colpali train set & 2,000 & Text-Image → Text \\
        VisualNews-t2i & 1,000 & Text → Text-Image \\
        VisualNews-i2t & 1,000 & Text-Image → Text \\
        VisRAG-Ret-Train-In-domain-data & 2,000 & Text → Image \\
        CIRR & 2,000 & Text-Image → Text-Image \\
        ChartQA & 1,000 & Text-Image → Text \\
        OK-VQA & 1,000 & Text-Image → Text \\
        llavahound qa & 2,000 & Video-Text → Text \\
        llavahound caption retrieval & 2,000 & Video → Text \\
        Visual7W & 1,000 & Text-Image → Text \\
        N24News & 1,000 & Text-Image → Text \\
        \midrule
        \textbf{Total} & 19,000 & Multimodal \\
        \bottomrule
    \end{tabular}
    }
\end{table}

\clearpage
\section{Detailed Scores of MMEB-V2}
\label{app:detail_score}
\definecolor{avgcolor}{RGB}{240,248,255}
\definecolor{icatcolor}{RGB}{255,245,235}
\definecolor{vcatcolor}{RGB}{240,255,240}
\definecolor{vdcatcolor}{RGB}{245,240,255}

We report the detailed metrics for our MMEB-V2 dataset~\cite{meng2025vlm2vec}, as shown in Table~\ref{tab:detailed_score_part1} and Table~\ref{tab:detailed_score_part2}. The highest and second-highest values are highlighted in \textbf{bold} and \underline{underlined}, respectively.

\begin{table}[htbp]
    \centering
    \caption{Detailed results of baselines and Embed-RL on full MMEB-v2 benchmark. Video and Visual Doc results are shown on the next table. }
    \renewcommand{\arraystretch}{1.1}
    \begin{adjustbox}{width=\textwidth}
    \begin{tabular}{l|cccccccccc}
        \toprule
        ~ & ColPali v1.3~\cite{faysse2024colpali} & GME-7B~\cite{zhang2025bridging} & VLM2Vec-7B~\cite{jiang2024vlm2vec} & VLM2Vec-V2-2B~\cite{meng2025vlm2vec} & CAFe-7B~\cite{yu2025cafe} & UME-R1-2B~\cite{lan2025ume} & UME-R1-7B~\cite{lan2025ume} & Embed-RL-2B & Embed-RL-4B \\ 
        \midrule
        \rowcolor{avgcolor}
        Avg - All (78 tasks) & 44.4 & 57.8 & 52.3 & 58.0 & 60.6 & 60.1 & 64.5 & \underline{66.8} & \textbf{68.1} \\ 
        \midrule
        \rowcolor{avgcolor}
        Avg - Image (36 tasks, Hit@1) & 34.9 & 56.0 & 65.5 & 64.9 & 67.6 & 66.6 & \textbf{71.3} & 69.2 & \underline{71.2} \\ 
        \rowcolor{avgcolor}
        Avg - Video (18 tasks, Hit@1) & 28.2 & 38.4 & 33.7 & 34.6 & 42.4 & 42.2 & 47.5 & \underline{52.1} & \textbf{53.0} \\ 
        \rowcolor{avgcolor}
        Avg - Visdoc (24 tasks, NDCG@5) & 71.0 & \textbf{75.2} & 46.4 & 65.4 & 63.9 & 63.9 & 67.1 & 74.1 & \underline{74.7} \\
        \midrule
        \rowcolor{icatcolor}
        I-CLS (10) & 40.3 & 57.7 & 62.7 & 62.9 & 63.6 & \underline{64.8} & \textbf{67.1} & 62.8 & 63.7 \\ 
        \rowcolor{icatcolor}
        I-QA (10) & 11.5 & 34.7 & 56.9 & 56.3 & 61.7 & 62.8 & \underline{69.2} & 67.9 & \textbf{70.5} \\ 
        \rowcolor{icatcolor}
        I-RET (12) & 48.1 & 71.2 & 69.4 & 69.5 & 69.1 & 67.6 & \textbf{71.9} & 68.6 & \underline{71.3} \\ 
        \rowcolor{icatcolor}
        I-VG (4) & 40.3 & 59.3 & 82.2 & 77.3 & 87.6 & 77.2 & 84.9 & \underline{90.4} & \textbf{91.4} \\ 
        \rowcolor{vcatcolor}
        V-CLS (5) & 26.7 & 37.4 & 39.1 & 39.3 & 35.8 & 44.3 & 48.6 & \underline{57.0} & \textbf{57.6} \\ 
        \rowcolor{vcatcolor}
        V-QA (5) & 37.8 & 50.4 & 30.0 & 34.3 & \underline{58.7} & 51.0 & \textbf{60.7} & 55.9 & 58.4 \\ 
        \rowcolor{vcatcolor}
        V-RET (5) & 21.6 & 28.4 & 29.0 & 28.8 & 34.4 & 32.9 & 38.2 & \textbf{45.1} & \textbf{45.1} \\ 
        \rowcolor{vcatcolor}
        V-MR (3) & 25.5 & 37.0 & 38.9 & 36.8 & 39.5 & 39.7 & 39.3 & \underline{49.4} & \textbf{49.5} \\ 
        \rowcolor{vdcatcolor}
        VD-Vidore-V1 (10) & \underline{83.6} & \textbf{89.4} & 56.9 & 75.7 & 70.7 & 72.4 & 75.7 & 79.9 & 80.2 \\ 
        \rowcolor{vdcatcolor}
        VD-Vidore-V2 (4) & 52.0 & \textbf{55.6} & 9.4 & 45.1 & 49.6 & 46.2 & 50.5 & 52.0 & \underline{53.4} \\ 
        \rowcolor{vdcatcolor}
        VD-VisRAG (6) & 81.1 & \textbf{85.0} & 59.1 & 79.6 & 79.5 & 79.2 & 83.7 & 84.6 & \underline{84.9} \\ 
        \rowcolor{vdcatcolor}
        VD-OOD (4) & 43.1 & 44.4 & 38.1 & 39.6 & 38.1 & 37.2 & 37.6 & \underline{65.7} & \textbf{67.1} \\
        \midrule

        ImageNet-1K & 42.4 & 64.6 & 80.1 & \textbf{80.8} & 77.3 & 75.3 & \underline{80.4} & 78.0 & 79.5 \\ 
        N24News & 25.5 & 50.5 & 79.7 & 72.9 & \textbf{83.2} & 81.1 & \underline{82.3} & 44.9 & 48.3 \\ 
        HatefulMemes & 50.6 & 53.6 & 69.7 & 56.3 & \underline{78.7} & 75.2 & \textbf{79.0} & 65.0 & 66.2 \\ 
        VOC2007 & 69.8 & 80.3 & 80.7 & 85.0 & \underline{89.8} & 80.0 & \textbf{90.8} & 78.7 & 79.5 \\ 
        SUN397 & 56.1 & 69.5 & 77.4 & 71.0 & \underline{79.9} & 79.4 & \textbf{80.3} & 75.4 & 79.2 \\ 
        Place365 & 27.5 & 39.1 & 37.4 & 35.9 & \underline{45.0} & 42.6 & \textbf{46.8} & 43.9 & 43.1 \\ 
        ImageNet-A & 14.9 & 41.2 & \underline{58.1} & 47.4 & 55.2 & 50.4 & 53.9 & \textbf{59.2} & \underline{58.1} \\ 
        ImageNet-R & 64.6 & 83.9 & 73.9 & \underline{89.3} & 88.0 & 88.7 & \textbf{90.1} & 88.5 & 88.2 \\ 
        ObjectNet & 45.6 & 69.0 & 40.1 & 65.2 & 22.5 & 52.0 & 42.3 & \underline{74.8} & \textbf{75.4} \\ 
        Country211 & 6.0 & 24.8 & \textbf{29.8} & \underline{25.2} & 16.7 & 23.4 & 25.0 & 20.0 & 19.4 \\ 
        OK-VQA & 9.4 & 33.2 & 56.8 & 51.5 & \underline{67.3} & 62.4 & \textbf{71.7} & 61.4 & \underline{67.3} \\ 
        A-OKVQA & 6.6 & 21.0 & 47.3 & 43.6 & \textbf{63.8} & 51.1 & 58.7 & 54.7 & \underline{59.3} \\ 
        DocVQA & 11.3 & 41.4 & 89.7 & 90.1 & 79.2 & 92.2 & \underline{93.8} & 92.4 & \textbf{94.3} \\ 
        InfographicsVQA & 5.0 & 20.3 & 60.0 & 58.8 & 53.3 & 67.7 & \textbf{79.2} & 76.7 & \underline{77.5} \\ 
        ChartQA & 5.7 & 17.8 & 56.9 & 47.4 & 48.8 & 64.9 & 75.1 & \underline{80.7} & \textbf{80.9} \\ 
        Visual7W & 6.1 & 22.2 & 52.7 & 52.9 & 52.5 & 54.1 & \underline{55.2} & 52.7 & \textbf{55.3} \\ 
        ScienceQA & 16.3 & 28.0 & 38.5 & 38.2 & \textbf{65.4} & 42.7 & 53.7 & 57.3 & \underline{61.6} \\ 
        VizWiz & 27.6 & 39.0 & 39.9 & 43.3 & 43.8 & 46.8 & 51.6 & \underline{54.5} & \textbf{56.2} \\ 
        GQA & 8.3 & \textbf{76.9} & 55.1 & 64.9 & 65.7 & 67.3 & \underline{69.3} & 64.9 & 68.5 \\ 
        TextVQA & 18.8 & 46.8 & 71.6 & 72.2 & 76.8 & 78.6 & 83.5 & \underline{83.8} & \textbf{84.3} \\ 
        VisDial & 41.2 & 60.8 & 81.9 & \underline{82.7} & \underline{82.7} & 76.6 & 80.7 & 81.5 & \textbf{84.9} \\ 
        CIRR & 8.2 & 54.9 & 51.1 & 57.5 & \underline{60.4} & 53.7 & 55.3 & 47.6 & \textbf{61.2} \\ 
        VisualNews\_t2i & 50.1 & \underline{79.7} & \textbf{80.5} & 74.5 & 69.5 & 71.7 & 76.8 & 71.9 & 73.7 \\ 
        VisualNews\_i2t & 47.6 & \textbf{83.6} & 81.2 & 78.2 & 79.4 & 74.2 & \underline{82.0} & 73.6 & 73.9 \\ 
        MSCOCO\_t2i & 59.2 & 71.2 & 77.2 & 75.3 & 75.4 & 75.1 & 78.3 & \textbf{79.4} & \underline{78.9} \\ 
        MSCOCO\_i2t & 49.9 & 57.7 & 73.9 & 71.4 & 73.1 & 68.9 & 71.4 & \underline{75.3} & \textbf{76.3} \\ 
        NIGHTS & 65.5 & 67.6 & 67.6 & \textbf{68.6} & 66.7 & 67.2 & \underline{68.1} & 66.3 & 66.4 \\ 
        WebQA & 53.8 & \textbf{91.4} & 88.3 & 90.6 & 89.3 & 90.0 & \underline{90.9} & 89.3 & 90.5 \\ 
        FashionIQ & 5.9 & \underline{37.8} & 17.1 & 19.5 & \textbf{39.0} & 17.1 & 23.4 & 24.0 & 31.9 \\ 
        Wiki-SS-NQ & \textbf{80.5} & \underline{78.2} & 62.3 & 66.9 & 61.2 & 62.0 & 72.5 & 68.9 & 69.6 \\ 
        OVEN & 50.0 & \textbf{75.1} & 66.5 & 64.3 & 60.8 & 66.9 & \underline{71.4} & 61.4 & 60.7 \\ 
        EDIS & 64.7 & \textbf{96.0} & 85.7 & 84.1 & 71.3 & 88.0 & \underline{92.0} & 84.5 & 87.4 \\ 
        MSCOCO & 36.7 & 31.4 & 75.7 & 67.1 & 84.7 & 69.5 & 72.7 & \underline{92.9} & \textbf{93.6} \\ 
        RefCOCO & 64.5 & 60.9 & 87.6 & 87.1 & 89.4 & 83.3 & 91.4 & \underline{94.9} & \textbf{95.9} \\ 
        RefCOCO-Matching & 3.9 & 78.4 & 84.6 & 85.8 & 83.0 & 84.4 & \textbf{91.1} & 85.8 & \underline{88.0} \\ 
        Visual7W-Pointing & 56.1 & 66.5 & 81.0 & 69.2 & \textbf{93.2} & 71.5 & 84.2 & \underline{88.0} & 87.9 \\ 
        \bottomrule
    \end{tabular}
    \end{adjustbox}
\label{tab:detailed_score_part1}
\end{table}

\begin{table}[htbp]
    \centering
    \caption{Detailed results of baselines and Embed-RL on Video and Visual Doc of MMEB-v2 benchmark.}
    \renewcommand{\arraystretch}{1.1}
    \begin{adjustbox}{width=\textwidth}
    \begin{tabular}{l|cccccccccc}
        \toprule
        ~ & ColPali v1.3~\cite{faysse2024colpali} & GME-7B~\cite{zhang2025bridging} & VLM2Vec-7B~\cite{jiang2024vlm2vec} & VLM2Vec-V2-2B~\cite{meng2025vlm2vec} & CAFe-7B~\cite{yu2025cafe} & UME-R1-2B~\cite{lan2025ume} & UME-R1-7B~\cite{lan2025ume} & Embed-RL-2B & Embed-RL-4B \\ 
        \midrule

        K700 & 23.4 & 39.7 & 35.5 & 38.0 & 40.1 & 35.8 & 42.8 & \underline{55.8} & \textbf{56.8} \\ 
        SmthSmthV2 & 25.1 & 30.6 & 32.1 & 42.8 & 35.8 & 44.1 & 50.4 & \underline{56.7} & \textbf{59.5} \\ 
        HMDB51 & 24.8 & 47.9 & 42.2 & 40.9 & 46.9 & 54.4 & \underline{58.3} & 56.7 & \textbf{60.1} \\ 
        UCF101 & 49.4 & 54.7 & 61.8 & 60.0 & 39.6 & 67.2 & 70.0 & \textbf{79.3} & \underline{78.5} \\ 
        Breakfast & 10.9 & 14.3 & 23.8 & 14.8 & 16.6 & 20.1 & 21.5 & \textbf{36.7} & \underline{33.0} \\ 
        MVBench & 33.7 & 46.6 & 28.5 & 33.7 & 48.9 & 49.9 & \textbf{58.2} & 50.8 & \underline{55.9} \\ 
        Video-MME & 30.6 & 39.2 & 27.8 & 30.7 & 46.0 & 41.7 & \underline{47.3} & 47.1 & \textbf{50.5} \\ 
        NExTQA & 35.2 & 53.6 & 20.3 & 20.9 & \underline{62.4} & 59.9 & \textbf{69.6} & 53.9 & 58.2 \\ 
        EgoSchema & 38.4 & 46.8 & 21.8 & 34.0 & \textbf{60.0} & 45.4 & 52.4 & \underline{53.0} & 52.8 \\ 
        ActivityNetQA & 51.3 & 65.6 & 51.4 & 52.3 & \textbf{76.0} & 57.8 & \textbf{76.0} & 74.8 & 74.4 \\ 
        DiDeMo & 22.8 & 26.4 & 29.3 & 30.4 & 37.8 & 32.4 & 40.0 & \underline{45.3} & \textbf{46.8} \\ 
        MSR-VTT & 17.6 & 31.8 & 34.5 & 28.3 & 36.5 & 34.3 & 38.9 & \underline{45.7} & \textbf{46.2} \\ 
        MSVD & 45.4 & 49.7 & 46.7 & 48.1 & 56.4 & 55.4 & 60.8 & \textbf{67.2} & \underline{65.8} \\ 
        VATEX & 16.7 & 24.9 & 25.5 & 26.5 & 32.0 & 29.9 & 32.6 & \textbf{43.6} & \underline{43.4} \\ 
        YouCook2 & 5.3 & 9.1 & 9.0 & 10.6 & 9.5 & 12.7 & 18.5 & \textbf{23.5} & \underline{23.3} \\ 
        QVHighlight & 19.9 & 59.5 & 57.7 & 49.4 & 58.4 & 57.5 & 54.9 & \underline{70.7} & \textbf{73.6} \\ 
        Charades-STA & \textbf{29.0} & 14.0 & 19.8 & 20.2 & 18.7 & 20.4 & 21.9 & \underline{26.4} & 25.0 \\ 
        MomentSeeker & 27.6 & 37.4 & 39.3 & 40.8 & 41.4 & 41.2 & 41.1 & \textbf{50.9} & \underline{49.9} \\ 
        \midrule

        ViDoRe\_arxivqa & 81.7 & \underline{86.9} & 60.2 & 80.6 & 73.3 & 73.9 & 73.6 & 86.1 & \textbf{88.7} \\ 
        ViDoRe\_docvqa & \underline{56.6} & \textbf{57.5} & 34.7 & 44.9 & 38.3 & 37.9 & 41.1 & 45.7 & 47.5 \\ 
        ViDoRe\_infovqa & 84.9 & \textbf{91.6} & 70.4 & 83.7 & 80.6 & 76.2 & 80.8 & 86.8 & \underline{86.9} \\ 
        ViDoRe\_tabfquad & 86.9 & \underline{94.6} & 78.2 & 89.2 & 80.7 & 86.1 & 90.2 & 94.5 & \textbf{94.7} \\ 
        ViDoRe\_tatdqa & \underline{70.9} & \textbf{74.1} & 27.6 & 43.8 & 37.8 & 40.6 & 46.7 & 54.6 & 54.8 \\ 
        ViDoRe\_shiftproject & \underline{75.1} & \textbf{96.8} & 38.6 & 60.8 & 52.0 & 66.8 & 65.0 & 70.7 & 69.0 \\ 
        ViDoRe\_artificial\_intelligence & \underline{95.7} & \textbf{99.6} & 67.7 & 88.5 & 86.0 & 85.9 & 89.5 & 94.0 & 91.6 \\ 
        ViDoRe\_energy & \underline{94.7} & \textbf{95.3} & 60.4 & 86.5 & 84.8 & 83.3 & 85.7 & 86.7 & 88.1 \\ 
        ViDoRe\_government\_reports & \underline{93.6} & \textbf{98.8} & 61.8 & 85.0 & 85.0 & 82.6 & 89.8 & 89.0 & 90.7 \\ 
        ViDoRe\_healthcare\_industry & \underline{95.9} & \textbf{99.3} & 69.9 & 92.2 & 88.4 & 90.8 & 94.3 & 91.1 & 90.4 \\ 
        ViDoRe\_esg\_reports\_human\_labeled\_v2 & 51.3 & \textbf{63.4} & 6.8 & 45.6 & 50.7 & 50.2 & 50.4 & 56.9 & \underline{59.8} \\ 
        ViDoRe\_biomedical\_lectures\_v2\_multilingual & \textbf{54.7} & 49.5 & 5.1 & 44.3 & 50.9 & 46.2 & 50.7 & \underline{51.0} & 50.1 \\ 
        ViDoRe\_economics\_reports\_v2\_multilingual & 49.0 & 54.2 & 13.9 & 43.0 & \underline{54.3} & 45.7 & \textbf{57.8} & 53.0 & 53.9 \\ 
        ViDoRe\_esg\_reports\_v2\_multilingual & \underline{52.9} & \textbf{55.4} & 11.9 & 46.6 & 42.3 & 42.7 & 43.2 & 46.9 & 49.7 \\ 
        VisRAG\_ArxivQA & 80.9 & \textbf{87.4} & 52.6 & 76.9 & 74.0 & 74.3 & 80.5 & 84.9 & \underline{86.9} \\ 
        VisRAG\_ChartQA & 72.3 & 86.1 & 57.7 & 83.7 & 82.7 & 86.0 & 85.0 & \underline{88.3} & \textbf{88.5} \\ 
        VisRAG\_MP-DocVQA & 82.0 & \textbf{89.7} & 60.6 & \underline{88.1} & 75.1 & 75.6 & 83.4 & 79.1 & 79.3 \\ 
        VisRAG\_SlideVQA & 85.1 & \textbf{92.6} & 54.7 & 84.1 & 87.6 & 87.1 & 91.5 & 92.3 & \textbf{92.6} \\ 
        VisRAG\_InfoVQA & 83.5 & 88.6 & 66.0 & 82.3 & 87.9 & 84.4 & 89.2 & \textbf{90.0} & \underline{89.6} \\ 
        VisRAG\_PlotQA & \textbf{79.3} & \underline{76.5} & 62.7 & 75.9 & 69.4 & 68.0 & 72.7 & 73.0 & 72.4 \\ 
        ViDoSeek-page & 38.1 & 32.6 & 16.3 & 29.1 & 22.5 & 21.2 & 21.3 & \underline{82.0} & \textbf{84.4} \\ 
        ViDoSeek-doc & \underline{87.5} & \textbf{90.3} & 69.4 & 79.0 & 73.8 & 75.9 & 75.3 & 82.6 & 82.4 \\ 
        MMLongBench-page & 27.1 & 36.9 & 0.4 & 15.8 & 13.3 & 11.9 & 12.3 & \underline{47.7} & \textbf{51.0} \\ 
        MMLongBench-doc & \underline{80.4} & \textbf{85.2} & 28.8 & 63.0 & 42.6 & 39.7 & 41.3 & 50.3 & 50.7 \\ 
        \bottomrule
    \end{tabular}
    \end{adjustbox}
\label{tab:detailed_score_part2}
\end{table}

\clearpage
\section{Detailed Scores of MMEB-V1}

We also report our performance on MMEB-V1~\cite{jiang2024vlm2vec}, including both in-domain and out-of-domain performance, as shown in Table~\ref{tab:v1_results}.

\begin{table*}[htbp]
\centering
\caption{Results on the MMEB-V1 benchmark (consisting of 36 image embedding tasks). IND and OOD denote the in-distribution and out-of-distribution datasets, respectively. The highest and second-highest values are highlighted in \textbf{bold} and \underline{underline}.}

\vspace{1mm}
\renewcommand{\arraystretch}{1.0}
\resizebox{0.8\textwidth}{!}{
\begin{tabular}{lccccccc}
\toprule
\multirow{2}{*}{Model} & \multicolumn{4}{c}{Per Meta-Task Score} & \multicolumn{3}{c}{Average Score} \\ \cline{2-8} 
& Classification & VQA & Retrieval & Grounding & IND & OOD & Overall \\ \midrule
\# of Datasets & 10 & 10 & 12 & 4 & 20 & 16 & 36 \\ \midrule
\rowcolor[HTML]{EDEDED}
\multicolumn{8}{c}{\textit{Baseline Models}} \\ \midrule
CLIP~\cite{radford2021learning} & 42.8 & 9.1 & 53.0 & 51.8 & 37.1 & 38.7 & 37.8 \\
BLIP2~\cite{li2023blip} & 27.0 & 4.2 & 33.9 & 47.0 & 25.3 & 25.1 & 25.2 \\
SigLIP~\cite{zhai2023sigmoid} & 40.3 & 8.4 & 31.6 & 59.5 & 32.3 & 38.0 & 34.8 \\
OpenCLIP~\cite{cherti2023reproducible} & 47.8 & 10.9 & 52.3 & 53.3 & 39.3 & 40.2 & 39.7 \\
UniIR (BLIP$_{\text{FF}}$)~\cite{wei2024uniir} & 42.1 & 15.0 & 60.1 & 62.2 & 44.7 & 40.4 & 42.8 \\
UniIR (CLIP$_{\text{SF}}$)~\cite{wei2024uniir} & 44.3 & 16.2 & 61.8 & 65.3 & 47.1 & 41.7 & 44.7 \\
Magiclens~\cite{zhang2024magiclens} & 38.8 & 8.3 & 35.4 & 26.0 & 31.0 & 23.7 & 27.8 \\ \midrule
\rowcolor[HTML]{EDEDED}
\multicolumn{8}{c}{\textit{MLLM-based Baseline Models}} \\ \midrule
E5-V~\cite{jiang2024e5} & 21.8 & 4.9 & 11.5 & 19.0 & 14.9 & 11.5 & 13.3 \\
VLM2Vec-2B~\cite{jiang2024vlm2vec} & 59.0 & 49.4 & 65.4 & 73.4 & 66.0 & 52.6 & 60.1 \\
VLM2Vec-7B~\cite{jiang2024vlm2vec} & 62.6 & 57.8 & 69.9 & 81.7 & 72.2 & 57.8 & 65.8 \\
VLM2Vec-V2~\cite{meng2025vlm2vec} & 62.9 & 56.3 & 69.5 & 77.3 & 68.8 & 59.9 & 64.9 \\
MMRet-7B~\cite{zhou2025megapairs} & 56.0 & 57.4 & 69.9 & 83.6 & 68.0 & 59.1 & 64.1 \\
CAFe-V1-7B~\cite{yu2025cafe} & 65.2 & 65.6 & 70.0 & 91.2 & \underline{75.8} & 62.4 & 69.8 \\
CAFe-V2-7B~\cite{yu2025cafe} & 63.6 & 61.7 & 69.1 & 87.6 & 72.8 & 61.1 & 67.6 \\
mmE5-11B~\cite{chen2025mme5} & \textbf{67.6} & 62.8 & 70.9 & 89.7 & 72.3 & \underline{66.7} & 69.8 \\
LLaVE-2B~\cite{lan2025llave} & 62.1 & 60.2 & 65.2 & 84.9 & 69.4 & 59.8 & 65.2 \\
LLaVE-7B~\cite{lan2025llave} & 65.7 & 65.4 & 70.9 & \textbf{91.9} & 75.0 & 64.4 & 70.3 \\
UniME-4B~\cite{gu2025breaking} & 54.8 & 55.9 & 64.5 & 81.8 & 68.2 & 52.7 & 64.2 \\
UniME-7B~\cite{gu2025breaking} & 66.8 & 66.6 & 70.6 & 90.9 & 74.6 & 65.8 & 70.7 \\
UME-R1-2B~\cite{lan2025ume} & 64.8 & 62.8 & 67.6 & 77.2 & 71.5 & 60.4 & 66.6 \\
UME-R1-7B~\cite{lan2025ume} & \underline{67.1} & \underline{69.2} & \textbf{71.9} & 84.9 & \textbf{76.1} & 65.1 & \textbf{71.3} \\
\midrule
\rowcolor[HTML]{FFE5E5}
\multicolumn{8}{c}{\textit{Ours}} \\ \midrule
Embed-RL-2B & 62.8 & 67.9 & 68.6 & 90.4 & 71.9 & 65.9 & 69.2 \\
Embed-RL-4B & 63.7 & \textbf{70.5} & \underline{71.3} & \underline{91.4} & 74.3 & \textbf{67.3} & \underline{71.2} \\
\bottomrule
\end{tabular}
}
\label{tab:v1_results}

\vspace{-2mm}
\end{table*}

\clearpage
\section{Prompt for synthesizing multimodal chain of thought}
\label{app:prompt}

To enable precise guidance for visual reasoning and retrieval tasks, we design a hierarchically structured prompting system that instructs models to execute visual analysis tasks across text, image, and video modalities. This system consists of two scenario-specialized core modules, which impose constraints on reasoning logic, output formatting, and evidence anchoring, and adheres to a framework of two-round reasoning with fixed-format output to guarantee the consistency and accuracy of the generated results.

\subsection{Basic Visual Reasoning Prompts}
Such prompts guide the model to complete basic reasoning based on inputs, supporting various visual tasks, and are divided into 4 items according to inputs and objectives:
\begin{itemize}
\item \textbf{Text-to-Image Retrieval Reasoning Prompts}: Focus on text-to-image retrieval, extract key visual concepts, anchor textual evidence to output a JSON keyword list, adapt to subtasks, and output results in a fixed format.

\item \textbf{Image Reasoning Prompts}: For image-based tasks, anchor visual evidence to output the 2D bounding box coordinates of key elements, locate core features, and complete reasoning and answer output in accordance with the process.

\item \textbf{Video Reasoning Prompts}: Adapt to video sequence tasks, output 1-based key frame indices based on frame visual evidence, identify core frames, and generate results following a fixed process.

\item \textbf{Text-to-Video Retrieval Reasoning Prompts}: For text-to-video retrieval, extract visual concepts containing temporal dynamics, output a JSON keyword list, and standardize result output according to subtasks.
\end{itemize}

\begin{promptbox}{Text-to-Image Retrieval Visual Reasoning Prompt}
\begin{lstlisting}[style=promptstyle]
You are a visual reasoning assistant specialized in text-to-image retrieval. Given a text description and a task, analyze the text content to determine the key visual concepts needed for retrieving matching images.

**Rules:**
1.  Keep reasoning concise and grounded in textual evidence. Limit each step to 1-2 sentences.
2.  Base your reasoning solely on the textual content and the task description.
3.  Rephrase the final answer to preserve its exact meaning, changing only wording/phrasing if needed.
4.  In your thinking process, you must extract and output key visual concepts from the text description. Use JSON format with key 'text_keywords' to specify the important keywords as a list.
5.  First, think between `<thinking>` and `</thinking>` while output necessary keywords from the text in JSON with key 'text_keywords'. Then, based on the thinking contents, rethink between `<rethink>` and `</rethink>`. Finally, output the answer within `<answer>...</answer>`.
6.  Your thought process should adapt to the task type:
    *   For **caption-based retrieval** (e.g., find image from caption): Extract key visual elements, objects, scenes, and relationships.
    *   For **news retrieval** (e.g., find news image from headline): Identify key people, locations, events, and contextual elements.
    *   For **dialogue-based retrieval** (e.g., find image from conversation): Summarize visual attributes, actions, and scene details.
    *   For **question-based retrieval** (e.g., find factual image): Identify the key concepts and relationships.
Now, process the following input:
TASK/QUESTION: {question}
\end{lstlisting}
\end{promptbox}

\begin{promptbox}{Image-based Visual Reasoning Prompt}
\begin{lstlisting}[style=promptstyle]
You are a visual reasoning assistant. Given an image and a task description or question, analyze the image step-by-step to produce the required output. The task may involve image retrieval, classification, question answering, or object identification.

**Rules:**
1.  Keep reasoning concise and grounded in visual evidence. Limit each step to 1-2 sentences.
2.  Base your reasoning solely on the visual content of the image and the task description.
3.  Rephrase the final answer to preserve its exact meaning, changing only wording/phrasing if needed.
4.  In your thinking process, you must output coordinates for the key visual element(s) relevant to answering the question. Use JSON format with key 'bbox_2d' to specify the bounding box as [x1, y1, x2, y2]. For multiple elements, use a list of bboxes: [[x1, y1, x2, y2], [x1, y1, x2, y2]].
5.  First, think between `<thinking>` and `</thinking>` while output necessary coordinates needed to answer the question in JSON with key 'bbox_2d'. Then, based on the thinking contents and coordinates, rethink between `<rethink>` and `</rethink>`. Finally, output the answer within `<answer>...</answer>`.
6.  Your thought process should adapt to the task:
    *   For **retrieval** (e.g., find similar image): Identify and locate key visual elements that define the match.
    *   For **classification** (e.g., scene, object, domain): Locate distinguishing visual features that belong to the class.
    *   For **question answering**: Locate the visual or textual clues in the image that lead to the answer.
    *   For **object identification/segmentation**: Provide the object's location and boundaries.

Now, process the following input:
IMAGE: {image}
TASK/QUESTION: {question}
\end{lstlisting}
\end{promptbox}

\begin{promptbox}{Video Sequence Visual Reasoning Prompt}
\begin{lstlisting}[style=promptstyle]
You are a video reasoning assistant. Given a video sequence (multiple frames) and a task description, analyze the video content step-by-step to produce the required output. The task may involve video captioning, video question answering, or video retrieval.

**Rules:**
1.  Keep reasoning concise and grounded in visual evidence from the video frames. Limit each step to 1-2 sentences.
2.  Base your reasoning solely on the visual content of the video frames and the task description.
3.  Rephrase the final answer to preserve its exact meaning, changing only wording/phrasing if needed.
4.  In your thinking process, you must identify and output key frames from the video sequence that are most relevant to answering the question. Use JSON format with key 'key_frames' to specify the frame indices as a list (using 1-based indexing).
5.  First, think between `<thinking>` and `</thinking>` while output necessary key frame indices in JSON with key 'key_frames'. Then, based on the thinking contents, rethink between `<rethink>` and `</rethink>`. Finally, output the answer within `<answer>...</answer>`.
6.  Your thought process should adapt to the task type:
    *   For **video captioning/description**: Identify frames that show main events, transitions, or key moments in the video sequence.
    *   For **video question answering**: Locate frames that contain the visual evidence needed to answer the specific question.
    *   For **video retrieval**: Identify frames that represent the core content or distinguishing features of the video.
Now, process the following input:
VIDEO FRAMES: {video}
TASK/QUESTION: {question}
\end{lstlisting}
\end{promptbox}

\begin{promptbox}{Text-to-Video Retrieval Visual Reasoning Prompt}
\begin{lstlisting}[style=promptstyle]
You are a visual reasoning assistant specialized in text-to-video retrieval. Given a video description and a task, analyze the text content to determine the key visual concepts needed for retrieving matching video clips or keyframes.

**Rules:**
1.  Keep reasoning concise and grounded in textual evidence. Limit each step to 1-2 sentences.
2.  Base your reasoning solely on the textual content and the task description.
3.  Rephrase the final answer to preserve its exact meaning, changing only wording/phrasing if needed.
4.  In your thinking process, you must extract and output key visual concepts from the text description. Use JSON format with key 'text_keywords' to specify the important keywords as a list.
5.  First, think between `<thinking>` and `</thinking>` while output necessary keywords from the text in JSON with key 'text_keywords'. Then, based on the thinking contents, rethink between `<rethink>` and `</rethink>`. Finally, output the answer within `<answer>...</answer>`.
6.  Your thought process should adapt to the task type:
    *   For **video-based retrieval** (e.g., find video from description): Extract key visual elements, objects, scenes, actions, temporal sequences, and relationships.
    *   For **scene-based retrieval** (e.g., find video from scene description): Identify key people, locations, events, contextual elements, and temporal progression.
    *   For **action-based retrieval** (e.g., find video from action sequence): Summarize visual attributes, actions, scene details, and temporal dynamics.
Now, process the following input:
TASK/QUESTION: {question}
\end{lstlisting}
\end{promptbox}

\subsection{Positive Sample Verification Reasoning Prompts}
These prompts analyze the rationale behind positive samples to support model training and verification, and are categorized into three types according to the positive sample modality:
\begin{itemize}
\item \textbf{Text Positive Sample Analysis Prompts}: Anchor textual evidence to output a JSON-formatted keyword list, generate a text summary through two rounds of reasoning, and clarify the rationale for labeling the sample as positive.

\item \textbf{Image Positive Sample Analysis Prompts}: Infer image content in conjunction with the target task, output bounding box coordinates of key elements, generate an image summary through two rounds of reasoning, and clarify the core features and grounding basis.

\item \textbf{Video Positive Sample Analysis Prompts}: Anchor video evidence to output 1-indexed key frame indices, generate a structured video analysis through two rounds of reasoning, and clarify the core rationale for positive sample validity.
\end{itemize}

\begin{promptbox}{Positive Text Output Visual Reasoning Analysis Prompt}
\begin{lstlisting}[style=promptstyle]
You are a visual reasoning assistant. Given a task description and a positive text output, analyze the text to determine the key concepts that make it the correct output.

**Rules:**
1. Keep reasoning concise and grounded in the textual evidence. Limit each step to 1-2 sentences.
2. Use the task description to understand the context of the output.
3. In your thinking process, you must extract and output key concepts from the text. Use JSON format with key 'text_keywords' to specify the important keywords as a list.
4. First, think between `<thinking>` and `</thinking>` while output necessary keywords from the text in JSON with key 'text_keywords'. Then, based on the thinking contents, rethink between `<rethink>` and `</rethink>`. Finally, output a brief description of the text within `<answer>...</answer>`.
Now, process the following input:
POSITIVE TEXT OUTPUT: {pos_text}
\end{lstlisting}
\end{promptbox}

\begin{promptbox}{Positive Image Output Visual Reasoning Analysis Prompt}
\begin{lstlisting}[style=promptstyle]
You are a visual reasoning assistant. Given a task description and a positive image output (with optional text template), analyze the task to determine what the image should contain and locate key regions.

**Rules:**
1. Keep reasoning concise and grounded in the task description. Limit each step to 1-2 sentences.
2. Use the task description to infer what the target image should look like.
3. In your thinking process, you must output coordinates for the key visual element(s) relevant to answering the question. If no question is specified, output the most important element. Use JSON format with key 'bbox_2d' to specify the bounding box as [x1, y1, x2, y2]. For multiple elements, use a list of bboxes: [[x1, y1, x2, y2], [x1, y1, x2, y2]].
4. First, think between `<thinking>` and `</thinking>` while output necessary coordinates in JSON with key 'bbox_2d'. Then, based on the thinking contents, rethink between `<rethink>` and `</rethink>`. Finally, output a brief description of the image within `<answer>...</answer>`.
Now, process the following input:
POSITIVE TEXT OUTPUT: {pos_text}
POSITIVE IMAGE OUTPUT: {pos_image_description} 
\end{lstlisting}
\end{promptbox}

\begin{promptbox}{Positive Video Output Visual Reasoning Analysis Prompt}
\begin{lstlisting}[style=promptstyle]
You are a visual reasoning assistant for video. Given a task description and a positive video output, analyze the output to determine the key moments or concepts that make it correct for the task.

**Rules:**
1.  Keep reasoning concise and grounded in the provided video output evidence. Limit each step to 1-2 sentences.
2.  Use the task description to understand the context of the output.
3.  In your thinking process, you must identify and output the indices of the most relevant or representative frames from the described video sequence. Use JSON format with key 'key_frames' to specify the frame indices as a list (using 1-based indexing).
4.  First, think between `<thinking>` and `</thinking>` while outputting necessary key frame indices in JSON with key 'key_frames'. Then, based on the thinking contents, rethink between `<rethink>` and `</rethink>`. Finally, output a brief analysis of the video output within `<answer>...</answer>`.
Now, process the following input:
POSITIVE TEXT OUTPUT: {pos_text}
POSITIVE VIDEO OUTPUT: {pos_video_output}
\end{lstlisting}
\end{promptbox}

\clearpage
\section{Comprehensive Performance Characterization on Video Retrieval}
\label{app:uvrb}
To fully validate the cross-dimensional generalization capability of our proposed model in complex and diverse retrieval scenarios, we conduct a systematic performance evaluation on the Universal Video Retrieval Benchmark (UVRB)~\cite{guo2025towards}. This benchmark consists of 16 datasets targeting distinct core capabilities, comprehensively covering multiple retrieval paradigms (textual, composed, and visual retrieval) and diverse semantic scenarios (coarse-grained, fine-grained, and long-context retrieval). It thus enables accurate quantification of the model's universal adaptation capacity across heterogeneous retrieval tasks. Experimental results demonstrate that our model outperforms counterparts with equivalent parameter scales, maintains a consistent performance advantage, and achieves the optimal comprehensive capability among models of the same parameter scale. Specifically, in core retrieval dimensions: (1) it attains state-of-the-art performance in coarse-grained semantic retrieval tasks and the second-best result in fine-grained semantic understanding scenarios; (2) it secures the optimal and second-best performances in spatial fine-grained perception (object/appearance recognition) and temporal fine-grained perception (motion/dynamics capture) subtasks, respectively. These results fully highlight the model's robust capability in multi-dimensional semantic understanding and spatiotemporal feature extraction.

\subsection{Video Retrieval Performance on UVRB Datasets}

\begin{table}[htbp]
\centering
\footnotesize
\setlength{\tabcolsep}{2.5pt}
\caption{Performance of video retrieval on UVRB datasets: AVG values represent the average across 16 datasets, with the highest score in each column bolded and the second-highest underlined. Metrics include R@1 (Recall@1), R@10 (Recall@10) and P@1 (Precision@1). The highest and second-highest values are highlighted in \textbf{bold} and \underline{underline}.}
\label{tab:main_by_datasets}

\resizebox{\textwidth}{!}{
\begin{tabular}{l|>{\columncolor{gray!15}}c|*{8}{c}}
\toprule
\textbf{Model} & \textbf{AVG} & \textbf{MSRVTT} & \textbf{DiDeMo} & \textbf{CRB-G} & \textbf{CRB-S} & \textbf{VDC-O} & \textbf{CRB-T} & \textbf{CMRB} & \textbf{DREAM-E} \\
& & R@1 & R@1 & R@1 & R@1 & R@1 & R@1 & R@10 & R@1 \\
\midrule
CLIP4Clip~\cite{luo2022clip4clip} & 39.0 & 33.3 & 29.7 & 51.1 & 49.7 & 62.0 & 28.9 & 28.0 & 19.1 \\
ViCLIP~\cite{wang2023internvid} & 35.2 & 38.6 & 30.6 & 44.7 & 43.7 & 53.0 & 34.9 & 22.9 & 23.5 \\
VideoCLIP-XL~\cite{wang2024videoclip} & 49.1 & 44.3 & 40.3 & 82.8 & 83.9 & 73.5 & 48.7 & 27.4 & 26.3 \\
LanguageBind~\cite{zhu2024languagebind} & 48.7 & \underline{47.9} & 42.1 & 71.6 & 68.7 & 75.9 & 46.6 & 29.0 & 28.0 \\
InternVideo2-1B~\cite{wang2024internvideo2} & 40.4 & 44.9 & 40.4 & 58.6 & 56.8 & 64.4 & 47.0 & 35.5 & 24.2 \\
InternVideo2-6B~\cite{wang2024internvideo2} & 42.7 & \textbf{48.5} & 41.8 & 60.8 & 61.2 & 65.0 & 45.5 & 34.6 & 27.1 \\
GME-2B~\cite{zhang2025bridging} & 48.8 & 39.0 & 30.3 & 69.0 & 71.8 & 71.5 & 40.0 & 29.8 & 24.0 \\
Unite-2B~\cite{kong2025modality} & 48.0 & 36.7 & 29.8 & 69.9 & 72.3 & 72.7 & 40.9 & 28.4 & 22.3 \\
VLM2Vec-V2~\cite{meng2025vlm2vec} & 50.8 & 33.0 & 29.9 & 82.8 & 84.3 & 77.5 & 41.0 & 28.6 & 22.8 \\
BGE-VL~\cite{zhou2024megapairs} & 44.3 & 33.7 & 31.8 & 69.0 & 68.8 & 63.9 & 35.9 & 22.5 & 21.2 \\
UniME-7B~\cite{gu2025breaking} & 52.1 & 35.1 & 33.5 & 81.5 & 82.7 & 74.3 & 47.6 & 31.7 & 29.3 \\
B3-7B~\cite{thirukovalluru2025breaking} & 51.1 & 28.2 & 35.0 & 81.5 & 82.5 & 76.8 & 41.5 & 31.2 & 21.6 \\
GME-7B~\cite{zhang2025bridging} & 53.0 & 43.6 & 37.7 & 74.0 & 76.7 & 73.1 & 44.2 & 30.4 & 27.4 \\
Unite-7B~\cite{kong2025modality} & 53.8 & 43.9 & 38.6 & 79.8 & 80.4 & 75.3 & 47.2 & 35.1 & 27.9 \\
GVE-3B~\cite{guo2025towards} & 54.4 & 43.1 & 37.6 & 85.0 & 84.6 & 78.6 & 49.6 & 36.3 & 28.0 \\
GVE-7B~\cite{guo2025towards} & 57.3 & 46.4 & \underline{43.3} & 86.5 & 84.7 & 79.4 & \textbf{53.9} & \textbf{39.8} & 30.2 \\
\midrule
Embed-RL-2B & \underline{58.7} & 43.7 & 42.2 & \underline{91.4} & \underline{89.7} & \underline{84.6} & 49.4 & 36.5 & \underline{31.8} \\
Embed-RL-4B & \textbf{60.2} & 44.0 & \textbf{46.1} & \textbf{92.1} & \textbf{89.9} & \textbf{85.9} & \underline{53.8} & \underline{38.2} & \textbf{32.0} \\
\bottomrule
\end{tabular}
}
\vspace{5pt}

\resizebox{\textwidth}{!}{
\begin{tabular}{l|*{8}{c}}
\toprule
\textbf{Model} & \textbf{LoVR-TH} & \textbf{PEV-K} & \textbf{LoVR-V} & \textbf{VDC-D} & \textbf{MS-TI} & \textbf{MS-TV} & \textbf{MSRVTT-I2V} & \textbf{LoVR-C2V} \\
& R@10 & R@1 & R@1 & R@1 & P@1 & P@1 & R@1 & R@1 \\
\midrule
CLIP4Clip~\cite{luo2022clip4clip} & 33.8 & 17.9 & 36.0 & 56.6 & 17.3 & 18.3 & \textbf{92.4} & 50.3 \\
ViCLIP~\cite{wang2023internvid} & 20.2 & 7.5 & 23.0 & 39.5 & 28.3 & 24.3 & 84.6 & 43.3 \\
VideoCLIP-XL~\cite{wang2024videoclip} & 43.9 & 22.9 & 38.0 & 82.0 & 23.0 & 22.3 & 86.1 & 40.3 \\
LanguageBind~\cite{zhu2024languagebind} & 42.5 & 30.3 & 54.0 & 67.9 & 22.8 & 23.3 & 82.7 & 46.3 \\
InternVideo2-1B~\cite{wang2024internvideo2} & 29.8 & 2.6 & 28.0 & 48.5 & 26.5 & 23.0 & 79.4 & 36.8 \\
InternVideo2-6B~\cite{wang2024internvideo2} & 30.2 & 8.6 & 33.0 & 51.6 & 23.5 & 20.5 & 86.8 & 45.2 \\
GME-2B~\cite{zhang2025bridging} & 44.6 & 35.4 & 53.0 & 83.9 & \textbf{35.0} & \textbf{34.0} & 82.7 & 36.6 \\
Unite-2B~\cite{kong2025modality} & 44.5 & 35.5 & 57.0 & 79.2 & 25.0 & 23.3 & 86.3 & 44.5 \\
VLM2Vec-V2~\cite{meng2025vlm2vec} & 49.2 & 32.4 & 61.0 & 91.3 & 27.5 & 25.0 & 84.1 & 38.5 \\
BGE-VL~\cite{zhou2024megapairs} & 38.7 & 18.4 & 55.0 & 72.2 & 30.3 & 23.3 & 77.9 & 46.5 \\
UniME-7B~\cite{gu2025breaking} & 50.4 & 32.3 & 48.0 & 84.7 & 31.0 & 30.5 & 86.7 & \textbf{53.7} \\
B3-7B~\cite{thirukovalluru2025breaking} & 46.2 & 38.7 & 59.0 & 85.3 & 27.5 & 26.5 & 88.4 & 47.1 \\
GME-7B~\cite{zhang2025bridging} & 52.3 & 39.6 & 71.0 & 86.5 & \underline{34.8} & \underline{33.3} & 86.0 & 37.0 \\
Unite-7B~\cite{kong2025modality} & 55.5 & \textbf{44.0} & 62.0 & 87.1 & 27.8 & 23.0 & 88.3 & 44.8 \\
GVE-3B~\cite{guo2025towards} & 52.2 & 33.0 & 61.0 & 91.8 & 34.0 & 26.8 & 89.1 & 40.3 \\
GVE-7B~\cite{guo2025towards} & 54.2 & \underline{41.3} & 68.0 & \underline{94.8} & 34.3 & 28.0 & \underline{89.9} & 41.5 \\
\midrule
Embed-RL-2B & \underline{56.5} & 33.6 & \textbf{80.0} & 93.8 & 19.3 & 21.0 & 89.1 & \underline{51.4} \\
Embed-RL-4B & \textbf{57.9} & 31.9 & \underline{77.0} & \textbf{95.2} & 15.8 & 21.0 & 87.9 & 49.0 \\
\bottomrule
\end{tabular}
}
\end{table}

\textbf{Baselines.} Following the experimental settings of GVE~\cite{guo2025towards}, we evaluate 16 representative baselines spanning diverse architectures, parameter scales, and training data compositions. These baselines are categorized into two groups: (1) traditional CLIP-based embedding models, including \path{CLIP4Clip}~\cite{luo2022clip4clip}, \path{ViCLIP}\cite{wang2023internvid}, \path{VideoCLIP-XL}~\cite{wang2024videoclip}, \path{LanguageBind}~\cite{zhu2024languagebind}, and the \path{InternVideo2} (1B/6B)\cite{wang2024internvideo2}; (2) recent MLLM-based embedding models, including \path{GVE-2B/7B}~\cite{guo2025towards}, \path{GME-2B/7B}~\cite{zhang2025bridging}, \path{Unite-2B/7B}~\cite{kong2025modality}, \path{VLM2Vec-V2}~\cite{meng2025vlm2vec}, \path{BGE-VL}~\cite{zhou2024megapairs}, \path{UniME-7B}~\cite{gu2025breaking}, and \path{B3-7B}~\cite{thirukovalluru2025breaking}.

\textbf{Datasets.} We adopt the Universal Video Retrieval Benchmark (UVRB)~\cite{guo2025towards}, which assesses model universality via 16 test datasets targeting distinct core abilities. UVRB covers diverse retrieval scenarios, with datasets categorized as: coarse-grained retrieval (MSRVTT~\cite{xu2016msr}, DiDeMo~\cite{anne2017localizing}, CRB-G~\cite{xu2025carebench}); fine-grained retrieval (spatial: CRB-S~\cite{xu2025carebench}/VDC-O~\cite{chai2024auroracap}, temporal: CRB-T~\cite{xu2025carebench}/CMRB~\cite{lin2025towards}, partially relevant: DREAM-E~\cite{wang2024tarsier}/LoVR-Theme2Clip~\cite{cai2025lovr}/PEV-K~\cite{bolya2025perception}); long-context retrieval (LoVR-V~\cite{cai2025lovr}, VDC-D~\cite{chai2024auroracap}); and composed query retrieval (MS-TI/MS-TV adapted from MomentSeeker~\cite{yuan2025momentseeker}, MSRVTT-I2V~\cite{xu2016msr}, LoVR-C2V~\cite{cai2025lovr}). 

\textbf{Metrics.} Following GVE~\cite{guo2025towards}, we adopt Recall@1 (R@1) as the primary metric, which measures the accuracy of identifying the most relevant item. For challenging datasets with ambiguous queries (e.g., CMRB, LoVR-TH), we additionally report Recall@10 (R@10) to reflect performance on top-k retrieval. For MS-TI and MS-TV (containing multiple positive candidates), we use Precision@1 (P@1) as the key metric.

\textbf{Performance.} Table \ref{tab:main_by_datasets} presents the video retrieval performance of diverse models on the Universal Video Retrieval Benchmark (UVRB)~\cite{guo2025towards} datasets. Our Embed-RL-2B and Embed-RL-4B models outperform all baselines in the average score (AVG), with Embed-RL-4B achieving the highest AVG (0.602). Both models secure the first or second place on key datasets across different retrieval paradigms, including coarse-grained retrieval (DiDeMo~\cite{anne2017localizing}, CRB-G~\cite{xu2025carebench}) and fine-grained/long-context retrieval (CRB-S~\cite{xu2025carebench}, VDC-O/VDC-D~\cite{chai2024auroracap}), which verifies their superior multi-dimensional video retrieval capability.

\subsection{Capability Characterization of the UVRB Evaluation Metrics}

The Universal Video Retrieval Benchmark (UVRB) adopts unweighted arithmetic means for all metric calculations to ensure fair comparison across heterogeneous datasets, and comprehensively evaluates video retrieval models from three orthogonal dimensions (Tasks, Domains, Sub-domains). 

UVRB covers 16 datasets that are exhaustively partitioned into non-overlapping categories according to Tasks, Domains and Sub-domains, as detailed in Table~\ref{tab:dataset_partition}. This partition is the foundation for quantifying distinct model capabilities in different retrieval scenarios.

\begin{table}[htbp]
  \centering
\caption{Detailed Partition of Datasets in the Universal Video Retrieval Benchmark (UVRB) Across Tasks, Domains, and Sub-domains}
\label{tab:dataset_partition}

  \begin{tabular}{@{}p{2.5cm}@{}p{10cm}@{}}
    \toprule
    \textbf{Partition} & \textbf{Content} \\
    \midrule
    $\mathcal{D}_{\text{TXT}}$ & $\{\text{MSRVTT}, \text{DiDeMo}, \text{CRB-G}, \text{CRB-S}, \text{VDC-O}, \text{CRB-T}, \text{CMRB},$ \\
    & $\quad \text{DREAM-E}, \text{LoVR-TH}, \text{PEV-K}, \text{LoVR-V}, \text{VDC-D}\}$ \\
    $\mathcal{D}_{\text{CMP}}$ & $\{\text{MS-TI}, \text{MS-TV}\}$ \\
    $\mathcal{D}_{\text{VIS}}$ & $\{\text{MSRVTT-I2V}, \text{LoVR-C2V}\}$ \\
    $\mathcal{D}_{\text{CG}}$  & $\{\text{MSRVTT}, \text{DiDeMo}, \text{CRB-G}\}$ \\
    $\mathcal{D}_{\text{FG}}$  & $\{\text{CRB-S}, \text{VDC-O}, \text{CRB-T}, \text{CMRB}, \text{DREAM-E}, \text{LoVR-TH}, \text{PEV-K}\}$ \\
    $\mathcal{D}_{\text{LC}}$  & $\{\text{LoVR-V}, \text{VDC-D}\}$ \\
    $\mathcal{D}_{\text{S}}$   & $\{\text{CRB-S}, \text{VDC-O}\}$ \\
    $\mathcal{D}_{\text{T}}$   & $\{\text{CRB-T}, \text{CMRB}\}$ \\
    $\mathcal{D}_{\text{PR}}$  & $\{\text{DREAM-E}, \text{LoVR-TH}, \text{PEV-K}\}$ \\
    \bottomrule
  \end{tabular}
\end{table}

Based on the above dataset partition, Table~\ref{tab:calc_rules} defines the calculation rules and corresponding capability characterization for each evaluation dimension. All metrics are computed as unweighted arithmetic means of corresponding datasets (denoted as $\mathcal{D}$), with results rounded to three decimal places. The three core dimensions are defined as follows:
\begin{enumerate}
    \item \textit{Task dimension}: Distinguishes retrieval paradigms by query formats (textual, composed, visual), reflecting cross-modal alignment ability for different query types;
    \item \textit{Domain dimension}: Assesses model performance across different levels of semantic granularity (coarse-grained, fine-grained, long-context), measuring generalization on retrieval tasks involving short/long context and high/low-level semantics;
    \item \textit{Sub-domain dimension}: Further decomposes fine-grained retrieval into three sub-tasks (spatial, temporal, partially relevant), pinpointing model strengths and weaknesses in fine-grained understanding.
\end{enumerate}

The overall AVG score is the arithmetic mean of the three task columns (TXT, CMP, VIS) and three domain columns (CG, FG, LC). This score aggregates model performance across core retrieval paradigms, rather than taking raw averages over datasets.

\begin{table}[htbp]
\centering
\small
\caption{Calculation Rules and Capability Characterization for Model Evaluation on UVRB (Table~\ref{tab:main_by_abilities}). All Metrics Are Computed as Unweighted Arithmetic Means of Corresponding Datasets (Denoted as $\mathcal{D}$), with Results Rounded to Three Decimal Places.}

\label{tab:calc_rules}
\resizebox{\textwidth}{!}{
\begin{tabular}{l l l}
\toprule
\textbf{Dimension Level} & \textbf{Column Name} & \textbf{Capability Characterization} \\
\midrule
Tasks & TXT & Text-to-video retrieval (12 datasets) \\
Tasks & CMP & Composed retrieval (text+image/video, 2 datasets) \\
Tasks & VIS & Visual-to-video retrieval (image/video, 2 datasets) \\
\midrule
Domains & CG & Coarse-grained semantic retrieval (3 datasets) \\
Domains & FG & Fine-grained semantic retrieval (7 datasets) \\
Domains & LC & Long-context retrieval (long text/video, 2 datasets) \\
\midrule
Sub-domains & S & Spatial fine-grained (object/appearance, 2 datasets) \\
Sub-domains & T & Temporal fine-grained (motion/dynamics, 2 datasets) \\
Sub-domains & PR & Partially relevant retrieval (3 datasets) \\
\midrule
Overall & AVG & Aggregated task and domains performance (core retrieval paradigms) \\
\bottomrule
\end{tabular}
}
\end{table}

Table~\ref{tab:main_by_abilities} presents the video retrieval performance of mainstream models on UVRB, characterized by the above-defined abilities (Tasks, Domains, and Sub-domains). The AVG score is the arithmetic mean of performance across textual (TXT), composed (CMP), and visual (VIS) retrieval tasks, aggregating model performance across core retrieval paradigms (rather than raw averages over datasets). 

\begin{table}[htbp]
\centering
\footnotesize
\caption{Video Retrieval Performance of Models on UVRB by Ability Dimensions (Tasks, Domains, Sub-domains). 
The AVG score denotes the mean performance across textual (TXT), composed (CMP), and visual (VIS) retrieval tasks. 
Domains involve coarse-grained (CG), fine-grained (FG), and long-context (LC) retrieval, while sub-domains include spatial (S), temporal (T), and partially relevant (PR) retrieval. 
The highest and second-highest values are highlighted in \textbf{bold} and \underline{underline}.}
\label{tab:main_by_abilities}

\resizebox{\textwidth}{!}{
\begin{tabular}{l|c|ccc|ccc|ccc}
\toprule
\textbf{Model} & & \multicolumn{3}{c|}{\textbf{Tasks}} & \multicolumn{3}{c|}{\textbf{Domains}} & \multicolumn{3}{c}{\textbf{Sub-domains}} \\
\cmidrule(lr){2-2} \cmidrule(lr){3-5} \cmidrule(lr){6-8} \cmidrule(lr){9-11}
& \textbf{AVG} & \textbf{TXT} & \textbf{CMP} & \textbf{VIS} & \textbf{CG} & \textbf{FG} & \textbf{LC} & \textbf{S} & \textbf{T} & \textbf{PR} \\
\midrule
CLIP4Clip~\cite{luo2022clip4clip} & 41.6 & 40.1 & 17.8 & \textbf{71.4} & 38.0 & 36.0 & 46.3 & 55.9 & 28.5 & 23.6\\
ViCLIP~\cite{wang2023internvid} & 37.5 & 33.6 & 26.3 & 64.0 & 38.0 & 31.5 & 31.3 & 48.4 & 28.9 & 17.1\\
VideoCLIP-XL~\cite{wang2024videoclip} & 51.0 & 55.0 & 22.7 & 63.2 & 55.8 & 49.3 & 60.0 & 78.7 & 38.1 & 31.0\\
LanguageBind~\cite{zhu2024languagebind} & 50.8 & 54.3 & 23.1 & 64.5 & 53.9 & 47.9 & 61.0 & 72.3 & 37.8 & 33.6\\
InternVideo2-1B~\cite{wang2024internvideo2} & 42.0 & 42.2 & 24.8 & 58.1 & 48.0 & 40.3 & 38.3 & 60.6 & 41.3 & 18.9\\
InternVideo2-6B~\cite{wang2024internvideo2} & 44.5 & 44.8 & 22.0 & 66.0 & 50.4 & 41.7 & 42.3 & 63.1 & 40.0 & 22.0\\
GME-2B~\cite{zhang2025bridging} & 41.6 & 53.9 & \textbf{34.5} & 59.7 & 46.1 & 47.1 & 68.5 & 71.6 & 34.9 & 34.7\\
Unite-2B~\cite{kong2025modality} & 50.7 & 53.6 & 24.2 & 65.4 & 45.5 & 47.1 & 68.1 & 72.5 & 34.7 & 34.1\\
VLM2Vec-V2~\cite{meng2025vlm2vec} & 53.8 & 58.7 & 26.3 & 61.3 & 49.8 & 50.2 & 76.2 & 80.9 & 34.8 & 34.8\\
BGE-VL~\cite{zhou2024megapairs} & 48.0 & 49.7 & 26.8 & 62.2 & 44.8 & 40.6 & 63.6 & 66.4 & 29.2 & 26.1\\
UniME-7B~\cite{gu2025breaking} & 54.2 & 56.1 & 30.8 & 70.2 & 50.0 & 51.8 & 66.4 & 78.5 & 39.6 & 37.3\\
B3-7B~\cite{thirukovalluru2025breaking} & 53.8 & 57.0 & 27.0 & 67.8 & 48.2 & 50.5 & 72.2 & 79.7 & 36.4 & 35.5\\
GME-7B~\cite{zhang2025bridging} & 56.2 & 60.4 & \underline{34.1} & 61.5 & 51.8 & 50.7 & 78.8 & 74.9 & 37.3 & 39.8\\
Unite-7B~\cite{kong2025modality} & 55.9 & 60.9 & 25.4 & 66.6 & 54.1 & 53.9 & 74.6 & 77.9 & 41.2 & \textbf{42.5}\\
GVE-3B~\cite{guo2025towards} & 57.1 & \underline{61.9} & 30.4 & 64.7 & 55.2 & 54.1 & 76.4 & 81.6 & \underline{43.0} & 37.7\\
\midrule
Embed-RL-2B & \textbf{58.7} & 61.1 & 20.1 & \underline{70.3} & \underline{59.1} & \underline{54.6} & \textbf{86.9} & \underline{87.2} & \underline{43.0} & \underline{40.6}\\
Embed-RL-4B & \underline{58.5} & \textbf{62.0} & 18.4 & \underline{70.3} & \textbf{60.7} & \textbf{55.6} & \underline{86.1} & \textbf{87.9} & \textbf{46.0} & \underline{40.6}\\

\bottomrule
\end{tabular}
}
\end{table}

\clearpage
\section{Training Trajectory Dynamics}\label{app:training_traj}

This section details dynamic training trajectories of 2B and 4B Embed-RL models via core metrics.

\subsection{Training Metrics}\label{app:training_metrics}

\begin{figure}[htbp]
    \centering
    \resizebox{\linewidth}{!}{\includegraphics{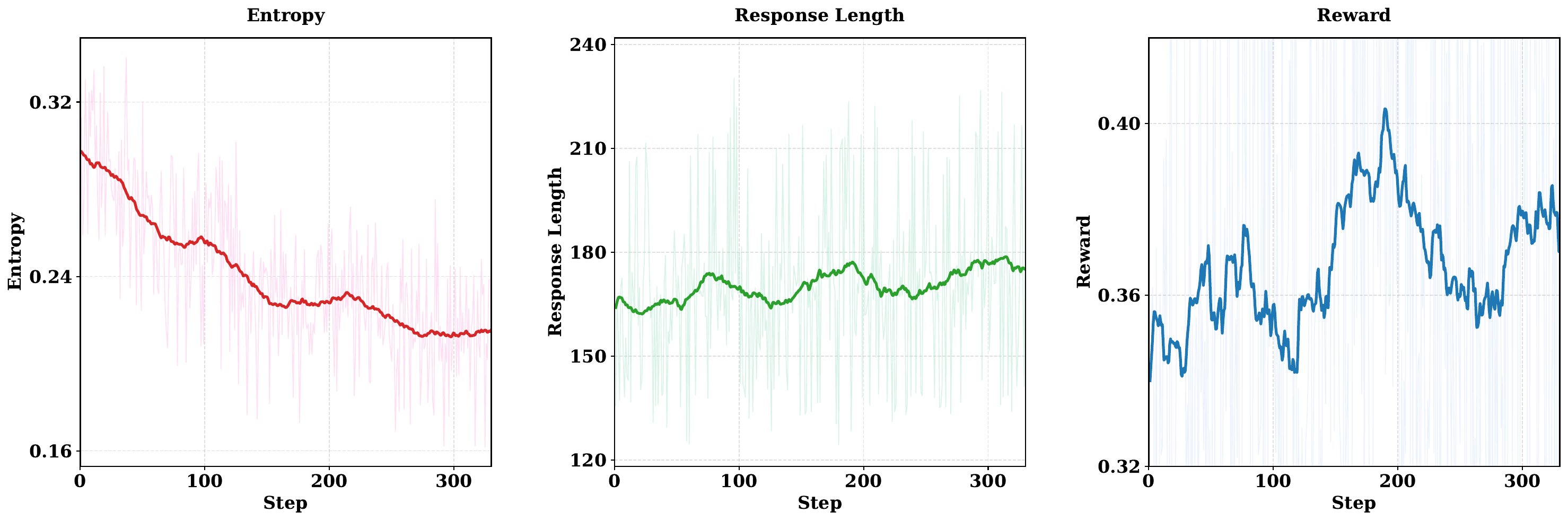}}
    \caption{Key RL-phase metrics of Embedder-Guided RL (entropy, response length, reward).}
    \label{fig:embed_rl_rl_metrics}
\end{figure}

 We track key metrics throughout the reinforcement learning phase, as shown in Figure~\ref{fig:embed_rl_rl_metrics}. We observe that the entropy declines gradually and then plateaus, while the response length increases steadily with ongoing training. Meanwhile, the reward exhibits a fluctuating upward trend, which is attributed to the effects of in-batch reward sample sampling as the discrepancies between individual samples far outweigh the inherent growth of the reward itself.

\begin{figure}[ht!]
    \centering
    \resizebox{\linewidth}{!}{\includegraphics{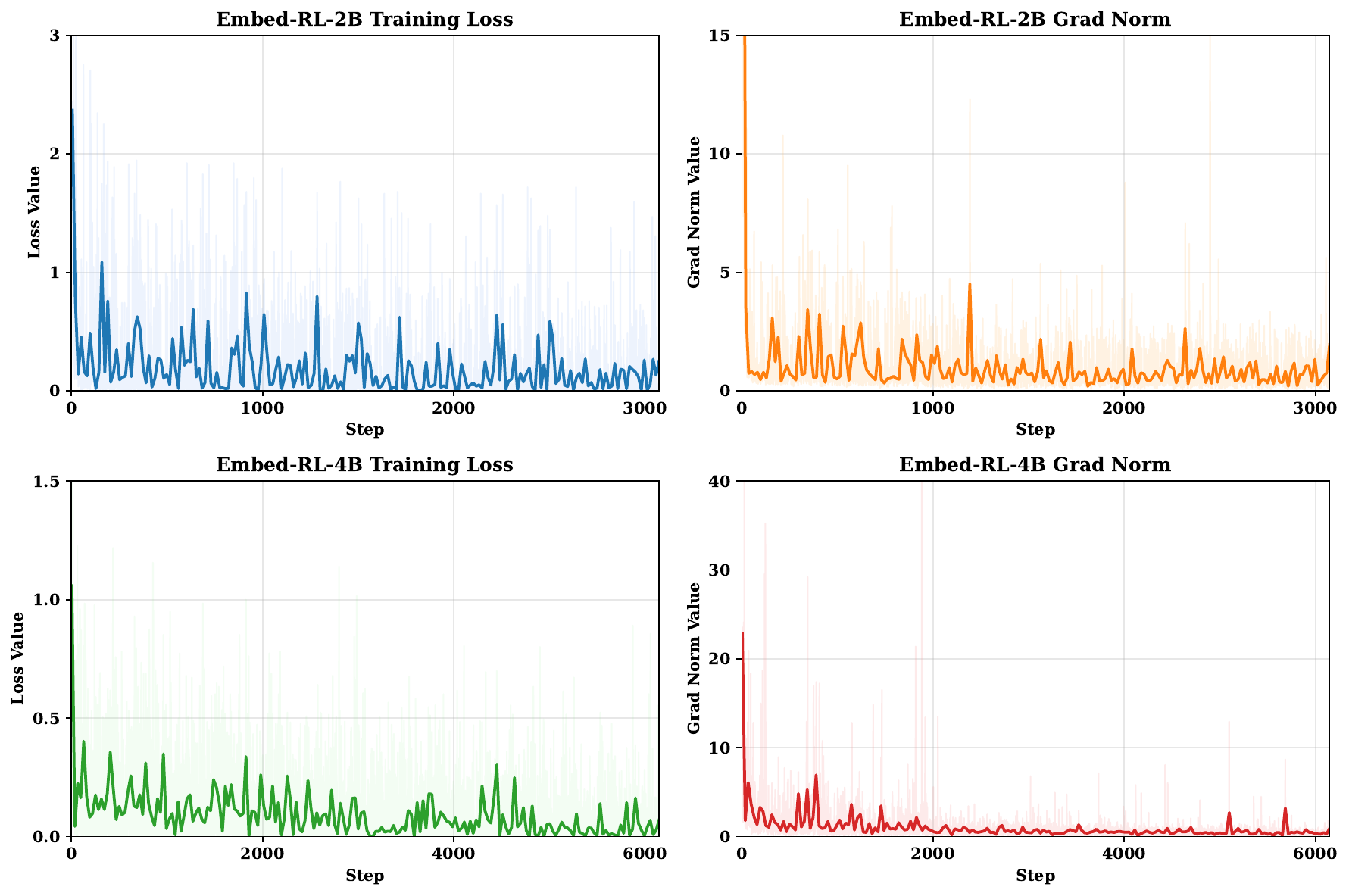}}
    \caption{Contrastive training loss and gradient norm of 2B and 4B scale Embed-RL models.}
    \label{fig:embed_rl_contrastive_metrics}
\end{figure}

Additionally, we track two core training metrics for 2B and 4B-scale Embed-RL models during contrastive learning: contrastive training loss and gradient norm, as presented in Figure~\ref{fig:embed_rl_contrastive_metrics}. Based on our training experience, neither an excessively large nor an overly small converged loss is favorable.An overly large converged loss suggests that the model fails to correctly discriminate positive and negative samples, whereas an excessively small one indicates that in-batch negative samples are too easily distinguished, leaving the model unable to learn effective discriminative information.Proper adjustment of the sampling ratio and sub-batch size is thus required to ensure the model converges correctly.

\clearpage

\section{Efficiency and Latency}
The proposed evidential Traceability CoT (T-CoT) demonstrates prominent efficiency advantages and negligible latency overhead compared with traditional generative embedding methods. For any multimodal retrieval target, T-CoT only needs to be generated once offline, and the derived multimodal embedding vector can be directly stored in the retrieval database. This is different from generative embedding approaches that require on-the-fly reasoning chain and embedding generation for each query, which leads to repeated computational costs. Designed to be targeted and concise, T-CoT only extracts core retrieval-related multimodal cues such as text keywords, image bounding boxes and video keyframes while abandoning redundant content. This ensures it does not significantly increase single embedding inference latency. Additionally, the stable semantic representation of T-CoT-based embeddings allows for long-term caching and reuse in subsequent tasks without frequent re-generation or updates. This further reduces inference latency and computational consumption in large-scale scenarios and makes the framework more suitable for practical industrial deployment.

\section{Limitations}
This work has several notable limitations. First, the weight coefficients of the multi-component reward function are empirically set for simplicity, lacking an adaptive optimization mechanism for diverse multimodal tasks, which may lead to suboptimal performance in specific scenarios. Second, the constructed dataset excludes partial classification tasks, resulting in relatively weak performance on image classification subtasks; we recommend designing additional loss for classification tasks to avoid false negatives while adapting to large-batch contrastive loss. Finally, we have not applied any hard negative sample mining or curriculum learning strategies, which are expected to further enhance the model’s discriminative capability and training stability if incorporated.

\section{Exploratory Perspectives}

In numerous practical systems such as Multimodal Content Understanding, Recommendation Systems (RS), and Retrieval-Augmented Generation (RAG), high-quality Universal Multimodal Embeddings serve as the core foundation supporting downstream tasks. The quality of embeddings directly determines the performance ceiling of downstream tasks—whether it is the matching accuracy of recommendation systems, the answer correctness of RAG systems, or the semantic alignment performance of cross-modal tasks, all are closely related to it.

In existing solutions, to extract richer semantic information, multi-dimensional feature fusion is often performed through techniques such as semantic segmentation, image and video fine-grained understanding, and user profile analysis. However, this approach typically incurs additional computational and storage overhead, and the feature fusion module suffers from high design complexity.

To address this, we strive to deeply internalize key information capture capabilities, including core visual region localization, video keyframe extraction, and core semantic keyword mining, into the embedding large model itself. This exploration aims to simplify the model structure of feature fusion, reduce the computational and storage costs of semantic vectors, and simultaneously enable embedding results to naturally incorporate task-relevant core semantic information.

\clearpage
\section{Comparative Examples of T-CoT Before and After Embedder-Guided RL}
\label{app:example}
\begin{figure}[htbp]
    \centering
    \includegraphics[width=\linewidth]{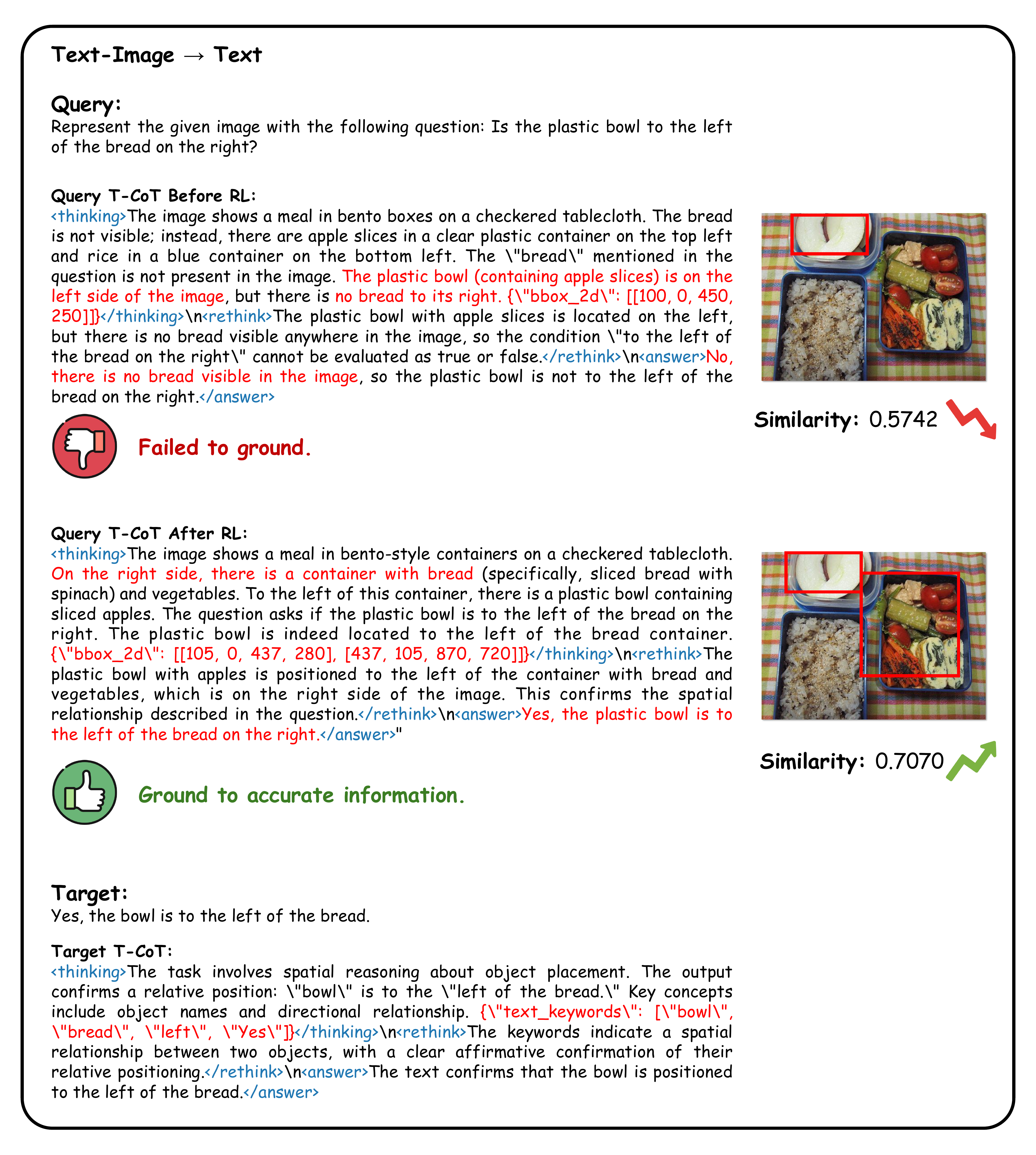}
    \caption{Example 4 of T-CoT Before and After EG-RL.}
    \label{fig:sup_example_1}
\end{figure}

Figures~\ref{fig:sup_example_1} to \ref{fig:sup_example_7} present additional comparative examples of T-CoT before and after the application of Embedder-Guided Reinforcement Learning. These examples fully demonstrate that after RL optimization, the Reasoner achieves more accurate localization of key regions and exhibits improved embedding quality, which further validates the effectiveness of the proposed EG-RL approach in enhancing the performance of T-CoT.

\begin{figure}[htbp]
    \centering
    \includegraphics[width=\linewidth]{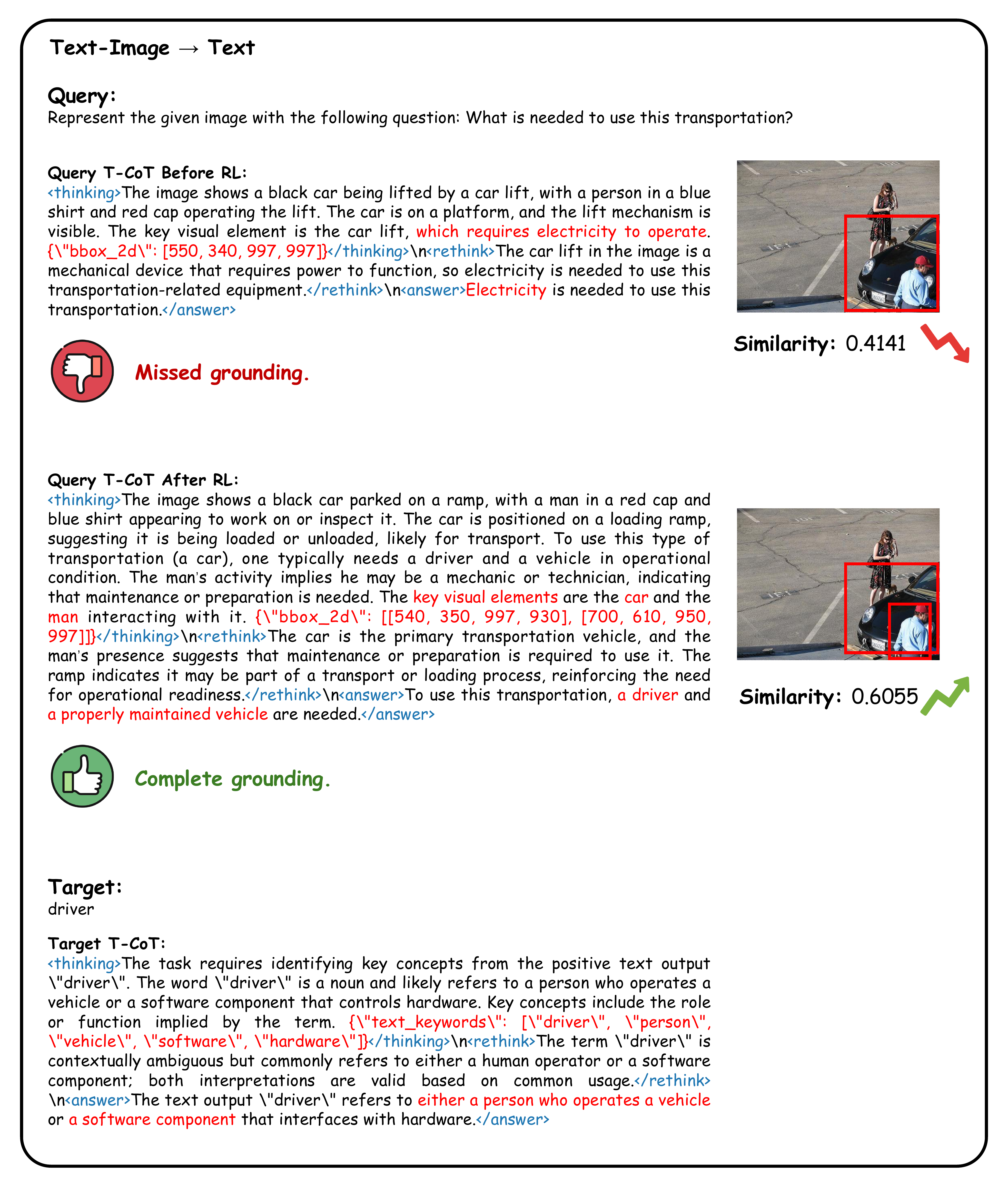}
    \caption{Example 5 of T-CoT Before and After EG-RL.}
    \label{fig:sup_example_3}
\end{figure}

\begin{figure}[htbp]
    \centering
    \includegraphics[width=\linewidth]{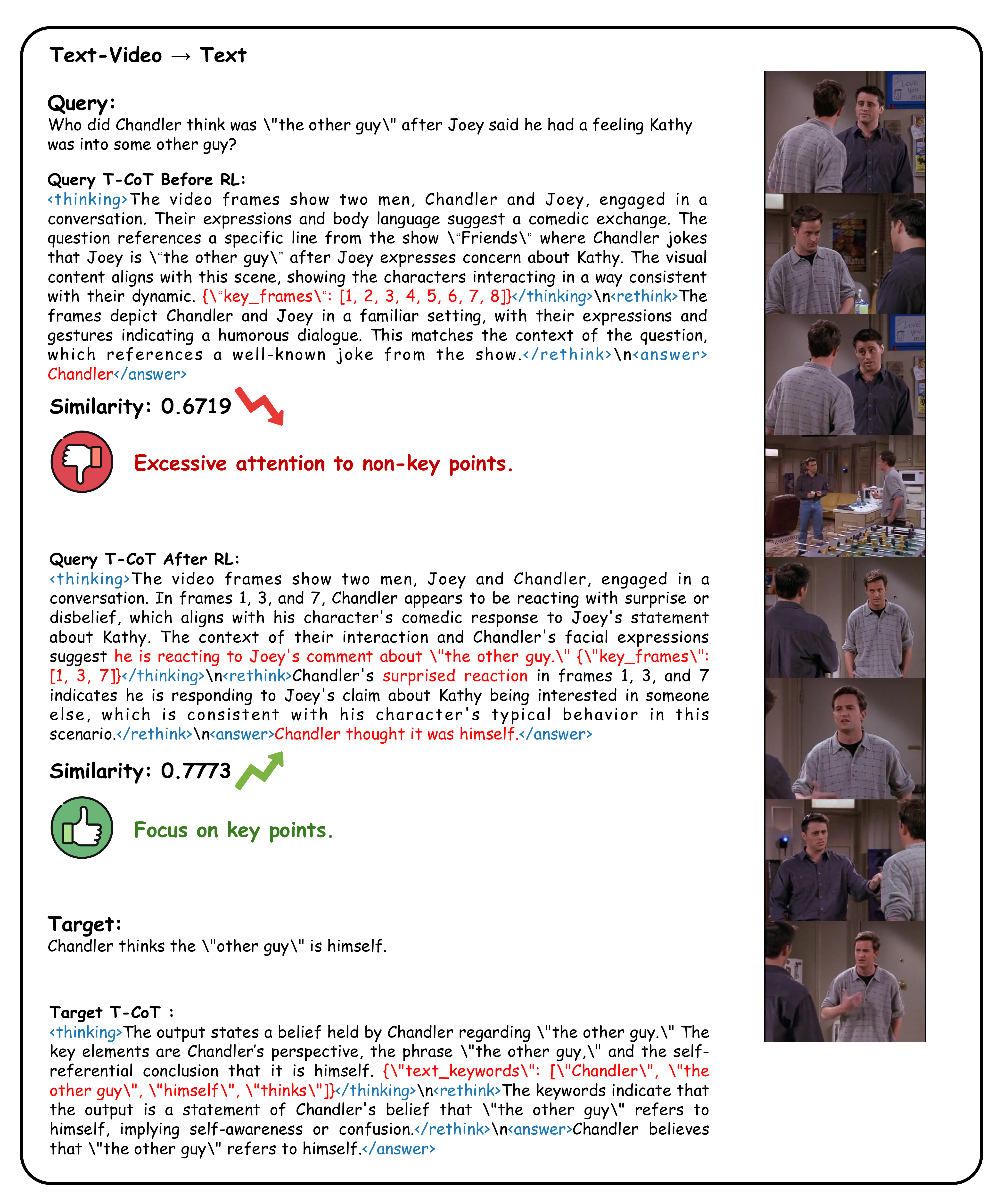}
    \caption{Example 6 of T-CoT Before and After EG-RL.}
    \label{fig:sup_example_5}
\end{figure}

\begin{figure}[htbp]
    \centering
    \includegraphics[width=\linewidth]{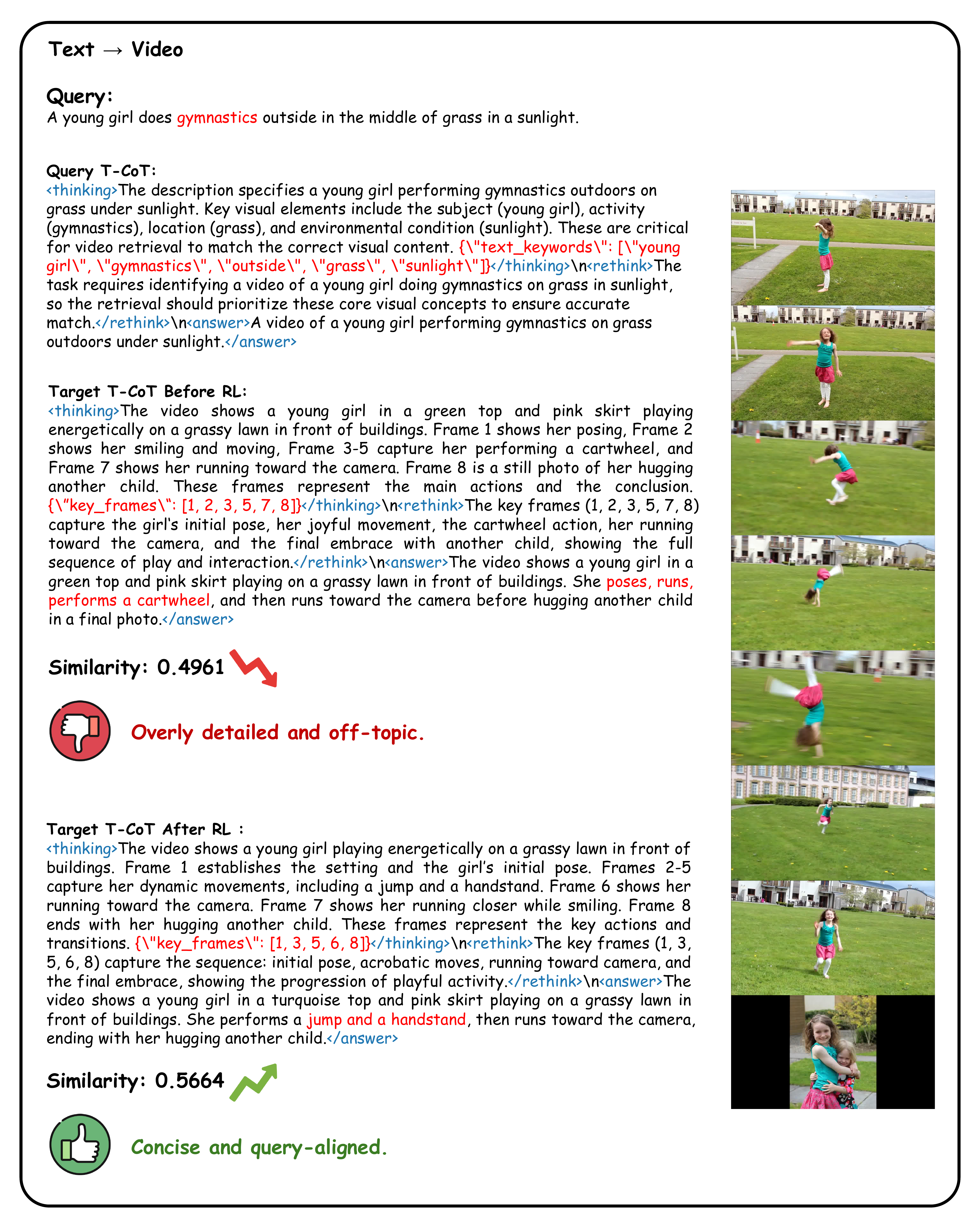}
    \caption{Example 7 of T-CoT Before and After EG-RL.}
    \label{fig:sup_example_6}
\end{figure}

\end{document}